\documentclass[11pt,a4paper]{article}

\usepackage[utf8]{inputenc}
\usepackage[T1]{fontenc}
\usepackage{lmodern}
\usepackage{amsmath,amssymb,amsfonts}
\usepackage{graphicx}
\graphicspath{{figures/}}
\usepackage{booktabs}
\usepackage{multirow}
\usepackage{caption}
\usepackage{subcaption}
\usepackage[colorlinks=true,linkcolor=blue!60!black,citecolor=blue!60!black,urlcolor=blue!60!black]{hyperref}
\usepackage{xcolor}
\usepackage{enumitem}
\usepackage[margin=1in]{geometry}
\usepackage{natbib}
\usepackage{float}
\usepackage{placeins}

\usepackage{titlesec}
\titlespacing*{\section}{0pt}{10pt plus 2pt minus 2pt}{5pt plus 1pt minus 1pt}
\titlespacing*{\subsection}{0pt}{8pt plus 2pt minus 2pt}{3pt plus 1pt minus 1pt}
\titlespacing*{\subsubsection}{0pt}{6pt plus 2pt minus 1pt}{2pt plus 1pt minus 1pt}

\title{Rethinking Uncertainty in Segmentation: \\ From Estimation to Decision}

\author{
  Saket Maganti
}

\date{}

\begin{document}

\maketitle

\begin{abstract}

In clinical practice, a segmentation model is only as useful as the
decision it supports. Yet most uncertainty research in medical imaging
stops at \emph{estimation}---producing variance maps and calibration
numbers---without saying what a clinician should actually \emph{do} with
them. That missing last step is what we study: the policy that converts
an uncertainty map into an action (accept, flag, or defer).

We recast uncertainty-aware segmentation as a two-stage pipeline,
estimation followed by decision, and show that the two stages interact
strongly enough that tuning either one alone leaves most of the available
safety gain on the table. On retinal vessel segmentation
(DRIVE, STARE, CHASE\_DB1), we pair Monte Carlo Dropout and Test-Time
Augmentation (TTA) with three deferral rules: a fixed global threshold,
an image-adaptive threshold, and a confidence-aware rule
$s = u \cdot (1-c)$ that we introduce to downweight pixels that are
uncertain but unambiguous.

Three findings emerge. First, the right pairing removes about 80\% of
segmentation errors at only 25\% pixel deferral, and the confidence-aware
rule removes 55\% of errors at just 12\% deferral---the best low-budget
operating point in our study. Second, TTA beats MC Dropout on both
decision quality (Unc-AUROC 0.881 vs.\ 0.722) and speed ($3.1\times$
faster), a practical result for deployment. Third, temperature
scaling---the standard calibration fix---does not improve deferral, and
uncertainty quality transfers surprisingly well under domain shift even
as Dice drops by 6.5 points. Together these results argue for a simple
evaluation principle: \textbf{judge uncertainty methods by the decisions
they enable, not by how calibrated their probabilities look in
isolation.}

\end{abstract}

\section{Introduction}
\label{sec:introduction}

A retinal vessel segmentation model that averages 0.78 Dice across a
test set tells a clinician very little about any single image. Some
images come back almost perfect; others contain errors that, if quietly
accepted, can change a screening decision for diabetic retinopathy or
glaucoma. The aggregate score hides this variance, and even a
well-calibrated probability map does not, on its own, tell a reader what
to \emph{do} next. What a real deployment needs is a \emph{policy}:
for every pixel, should the prediction be trusted, flagged for a second
look, or handed off to a human?

\paragraph{Estimation has outpaced decision.} Research on uncertainty in
medical image segmentation has concentrated almost entirely on
\emph{estimation}: better posteriors via Bayesian approximations
\citep{gal2016dropout}, ensembles \citep{lakshminarayanan2017simple}, and
test-time augmentation \citep{wang2019aleatoric}, scored by calibration
metrics such as ECE and NLL. These lines of work ask \emph{how well} we
estimate uncertainty; they rarely ask what happens next. Selective
prediction \citep{geifman2017selective,el2010foundations} formalizes
abstention in classification, but its application to dense pixel-level
prediction---and its coupling with modern uncertainty estimators---remains
underdeveloped. The result is a literature in which uncertainty maps are
produced, tabulated, and then left unused. The missing layer is
\emph{decision}: a policy that converts an uncertainty map into an action.
\textbf{Uncertainty without decisions is evaluation without consequence.}

\paragraph{A decision-aware perspective on uncertainty.} We reframe the
problem. The key question is not how well uncertainty is estimated, but
whether it enables better decisions. More pointedly: \textbf{uncertainty
is not a property of a model, but of a decision process.} Two models with
identical calibration can produce decisions of very different quality,
because the value of an uncertainty map depends on the rule that consumes
it. The same uncertainty map can look mediocre under one deferral rule
and excellent under another. These are not calibration statements; they
are statements about the joint estimation--decision system, and no
amount of recalibration addresses them.

This paper operationalizes that reframing. We study, empirically, how the
choice of uncertainty source and the choice of deferral policy jointly
determine segmentation decision quality, and we show that ignoring either
dimension leaves most of the available error reduction on the table. The
right combination of uncertainty method and deferral policy eliminates
roughly 80\% of segmentation errors while deferring only 25\% of pixels;
the wrong combination reduces errors by less than a third of that, using
the same images, the same model, and the same forward passes. The
difference is entirely in the \emph{decision layer} sitting on top.

\paragraph{Why this is not obvious.} One might object that ``use
uncertainty to defer'' is folklore, and therefore our reframing merely
restates what the field already knows. We disagree. In practice, existing
work optimizes uncertainty \emph{in isolation}: Bayesian methods are tuned
for calibration, ensembles for NLL, post-hoc methods for ECE. Deferral,
when it appears at all, is a fixed-threshold afterthought evaluated at a
single operating point. This decomposition hides the interaction we study.
A well-calibrated model with a poor deferral rule can underperform a
less-calibrated model with a well-designed one; any methodology that
scores uncertainty on its own is blind to this fact. The standard
calibration toolkit is not merely incomplete as a proxy for decision
quality---it is actively misleading, because it rewards distributional
match over discriminative separation, and deferral cares only about the
latter. Our contribution is to measure the interaction directly,
demonstrate that it is large, and show that the field has been optimizing
the wrong objective.

\paragraph{Setup.} Our study uses retinal vessel segmentation on DRIVE,
with cross-dataset evaluation on STARE and CHASE\_DB1. We train a U-Net
with a ResNet-34 encoder and compare two uncertainty estimation
approaches---MC Dropout (30 stochastic forward passes) and Test-Time
Augmentation (6 geometric transforms)---under three deferral strategies
of increasing sophistication:

\begin{enumerate}[leftmargin=*]
  \item \textbf{Global thresholding}: defer all pixels whose uncertainty
  exceeds a fixed threshold, optimized on a validation set.
  \item \textbf{Image-adaptive thresholding}: compute a per-image threshold
  based on percentile statistics of that image's uncertainty distribution,
  adapting the deferral rate to each image's difficulty.
  \item \textbf{Confidence-aware deferral}: weight uncertainty by inverse
  prediction confidence, so that uncertain pixels near the decision
  boundary ($p \approx 0.5$) are deferred preferentially over uncertain
  pixels where the model is nonetheless confident.
\end{enumerate}

We evaluate these combinations through risk-coverage curves, deferral
curves, and direct measurement of error reduction as a function of
deferral rate.

\paragraph{Contributions.} To our knowledge, this is the first study of
dense medical segmentation that jointly characterizes the uncertainty
source and the deferral policy as a single design space and measures the
interaction directly. Specifically:

\begin{enumerate}[leftmargin=*]
  \item \textbf{An estimation$\,\to\,$decision reframing of uncertainty
  for segmentation.} We treat estimation and deferral as one pipeline and
  show that their interaction is large enough that optimizing either
  alone leaves most of the available error reduction unrealized. Under
  this lens, ``uncertainty quality'' stops being a scalar property of a
  model and becomes a property of the decision rule it is paired with.

  \item \textbf{A confidence-aware deferral rule.} We introduce the
  score $s = u \cdot (1 - c)$, which downweights pixels that are
  uncertain but not borderline. Paired with TTA, it removes 55\% of
  errors at just 12\% deferral---the best low-budget operating point we
  observe. Paired with image-adaptive thresholding, the same uncertainty
  source reaches roughly 80\% error reduction at 25\% deferral, an
  upper bound on what the estimation signal can support.

  \item \textbf{TTA as the preferred decision-time uncertainty source.}
  A small, six-augmentation TTA ensemble beats a 30-pass MC Dropout
  ensemble on decision quality (Unc-AUROC 0.881 vs.\ 0.722) while
  running $3.1\times$ faster---a result that matters for deployment
  because the usual assumption is that more passes give better
  uncertainty.

  \item \textbf{Evidence that calibration is the wrong target for
  deferral.} Temperature scaling preserves deferral rankings and can
  even hurt ECE at the operating points we care about, confirming at the
  pixel level what prior classification work has shown: calibration and
  decision quality are genuinely decoupled and should be reported side
  by side.

  \item \textbf{Cross-dataset evidence that the decision layer is
  robust.} Zero-shot transfer to STARE and CHASE\_DB1 holds Unc-AUROC
  above 0.83 even as Dice drops by 6.5 points. The decision layer
  remains informative exactly when the estimation layer is most
  stressed---an encouraging property for real deployment across sites.
\end{enumerate}

The remainder of the paper proceeds as follows.
Sections~\ref{sec:related_work}--\ref{sec:experimental_setup} situate the
work, formalize the two-stage pipeline, and describe the experimental
protocol. Sections~\ref{sec:results}--\ref{sec:ablations} present results
and ablations. Sections~\ref{sec:discussion}--\ref{sec:conclusion} discuss
implications, limitations, and concluding arguments.

\section{Related work}
\label{sec:related_work}

\subsection{Medical image segmentation}

U-Net \citep{ronneberger2015unet} and its descendants remain the dominant
architecture family for medical image segmentation. Encoder-decoder designs
with skip connections have been extended in numerous directions: attention
gates \citep{oktay2018attention}, dense connectivity \citep{li2018hdenseunet},
transformer-based encoders \citep{chen2021transunet}, and multi-scale feature
fusion \citep{zhou2018unetplusplus}. For retinal vessel segmentation
specifically, the DRIVE \citep{staal2004drive}, STARE \citep{hoover2000stare},
and CHASE\_DB1 \citep{fraz2012chase} benchmarks have driven steady
improvements in Dice and AUC, with recent methods exceeding 0.80 Dice on DRIVE.

Nearly all of this work optimizes aggregate segmentation metrics. A model
that achieves 0.78 Dice is considered better than one at 0.76, regardless of
whether the errors cluster on ambiguous images (where a clinician would want
to look more carefully) or scatter randomly. This paper takes a different
starting point: given a segmentation model of reasonable quality, how should
its outputs be used in a decision pipeline?

\subsection{Uncertainty estimation in deep learning}

Bayesian neural networks provide a principled framework for uncertainty but are
computationally prohibitive for modern architectures \citep{neal1996bayesian,
blundell2015weight}. Practical approximations fall into three families.

\paragraph{MC Dropout.} Gal and Ghahramani \citep{gal2016dropout} showed that
applying dropout at test time and averaging multiple stochastic forward passes
approximates variational inference. The variance (or mutual information) across
passes quantifies epistemic uncertainty. MC Dropout is widely used in medical
imaging \citep{nair2020exploring,jungo2018effect,roy2019bayesian} because it
requires no architectural changes and works with any dropout-equipped network.
Its weakness is that the quality of the uncertainty estimate depends on
where dropout layers are placed and on the number of forward passes $T$, with
diminishing returns beyond $T \approx 20$--30 in our experiments. The
uncertainty also tends to be diffuse: MC Dropout flags large regions as
somewhat uncertain rather than precisely localizing errors.

\paragraph{Deep ensembles.} Lakshminarayanan et al.\
\citep{lakshminarayanan2017simple} train $N$ models with different random
initializations and use prediction variance across ensemble members as
uncertainty. Ensembles consistently produce better-calibrated predictions than
single models \citep{ovadia2019can}, but training $N$ separate models multiplies
computational cost. In our experiments, a 5-member ensemble requires roughly
$4\times$ the inference time of a single MC Dropout model.

\paragraph{Test-Time Augmentation.} TTA generates multiple predictions by
applying geometric or photometric transformations to the input and aggregating
the (inverse-transformed) outputs \citep{wang2019aleatoric,
ayhan2018test,moshkov2020tta}. Disagreement across augmented predictions
flags regions where the model's output is sensitive to input perturbations.
TTA captures a form of uncertainty that is complementary to MC Dropout:
where MC Dropout probes the model's weight posterior, TTA probes sensitivity
to input variations that should, in principle, not change the prediction. This
distinction is relevant because boundary regions and thin structures (common
error sources in vessel segmentation) tend to produce high TTA disagreement
even when MC Dropout uncertainty is moderate.

Mehrtash et al.\ \citep{mehrtash2020confidence} compared MC Dropout and TTA
for brain tumor segmentation and found that TTA-based uncertainty correlated
more strongly with segmentation error. Our work confirms this finding for
retinal vessels and extends it to the deferral setting, showing that TTA's
superior uncertainty quality translates directly to better deferral decisions.

\subsection{Calibration}

A model is well calibrated if its predicted probabilities match empirical
frequencies: among all pixels assigned probability 0.8, roughly 80\% should
be positive. Expected Calibration Error (ECE) \citep{naeini2015obtaining}
bins predictions by confidence and measures the average gap between predicted
and observed accuracy. Temperature scaling \citep{guo2017calibration} is the
standard post-hoc fix: a single scalar $T$ is learned on a validation set to
rescale logits before the sigmoid/softmax, shrinking or expanding the
confidence distribution.

The assumption behind calibration as a preprocessing step for decision-making
is that calibrated probabilities are more informative. This assumption is not
always correct. Calibration optimizes a global property (the probability
distribution matches reality on average) but deferral requires a local
property (high uncertainty should coincide with high error \emph{per pixel}).
A model can be perfectly calibrated yet have uncertainty that fails to
discriminate correct from incorrect predictions. We test this empirically
and find that temperature scaling barely moves ECE (from 0.038 to 0.038 for
MC Dropout, from 0.037 to 0.039 for TTA) and does not improve the tracked
deferral operating points.

Laves et al.\ \citep{laves2020well} reached a similar conclusion for
classification, observing that calibration and selective prediction can be
at odds. Our work extends this observation to dense prediction (segmentation)
and to the specific setting of pixel-level deferral.

\subsection{Selective prediction and deferral}

Selective prediction, also called prediction with a reject option, has a long
history in classification \citep{chow1970optimum,el2010foundations}. The idea
is simple: a model outputs both a prediction and a confidence score, and
predictions below a confidence threshold are withheld. Geifman and El-Yaniv
\citep{geifman2017selective} formalized risk-coverage curves for deep
classifiers, showing that selective prediction can dramatically reduce error
rates at modest coverage costs.

Translating selective prediction to segmentation is harder because the unit
of abstention is unclear. Should the model abstain on individual pixels, on
patches, or on entire images? Pixel-level deferral is the most granular but
produces fragmented deferral maps that may be impractical for human review.
Image-level deferral is cleaner but discards good predictions along with
bad ones. We adopt pixel-level deferral but evaluate it at multiple
granularities, including per-image error reduction and coverage-averaged
metrics.

DeVries and Taylor \citep{devries2018learning} trained a separate confidence
branch alongside a classifier. Corbiere et al.\ \citep{corbiere2019addressing}
proposed learning a failure prediction module. In medical imaging, Jungo et al.\
\citep{jungo2020analyzing} studied how uncertainty thresholds affect the
quality of retained predictions in brain lesion segmentation, and
Mehrtash et al.\ \citep{mehrtash2020confidence} evaluated deferral using
MC Dropout for prostate segmentation.

Our confidence-aware deferral strategy differs from these approaches in that
it does not require training an auxiliary network. Instead, it combines the
existing uncertainty map with the model's own prediction confidence through a
simple multiplicative score, $s = u \cdot (1 - c)$, where $u$ is uncertainty
and $c$ is prediction confidence. This has a clear interpretation: a pixel
is deferred when it is both uncertain \emph{and} the model's prediction is
near the decision boundary. Uncertain pixels where the model is nonetheless
confident (far from $p = 0.5$) are retained. This design costs nothing
computationally and, as we show, substantially improves deferral efficiency.

The learning-to-defer framework of Madras et al.\ \citep{madras2018predict}
and Mozannar and Sontag \citep{mozannar2020consistent} formalizes deferral as
a joint optimization of model and expert, assuming known expert accuracy. Our
setting is simpler and arguably more practical: we assume no model of the
human expert and instead optimize deferral thresholds on a validation set to
maximize a task-relevant objective (error reduction at minimum coverage cost).
This makes our approach applicable to any existing segmentation pipeline
without retraining.

\section{Methodology}
\label{sec:methodology}

Our framework treats segmentation as one stage in a two-stage decision pipeline.
The first stage produces a prediction and an uncertainty estimate. The second
stage, the deferral policy, consumes both and outputs a binary accept/defer
decision per pixel. The two stages are modular: any uncertainty method can be
paired with any deferral policy. This separation lets us isolate the effect of
each component and answer the question that matters for deployment: which
combination of uncertainty source and deferral rule produces the best decisions?

\begin{figure}[H]
\centering
\includegraphics[width=\textwidth]{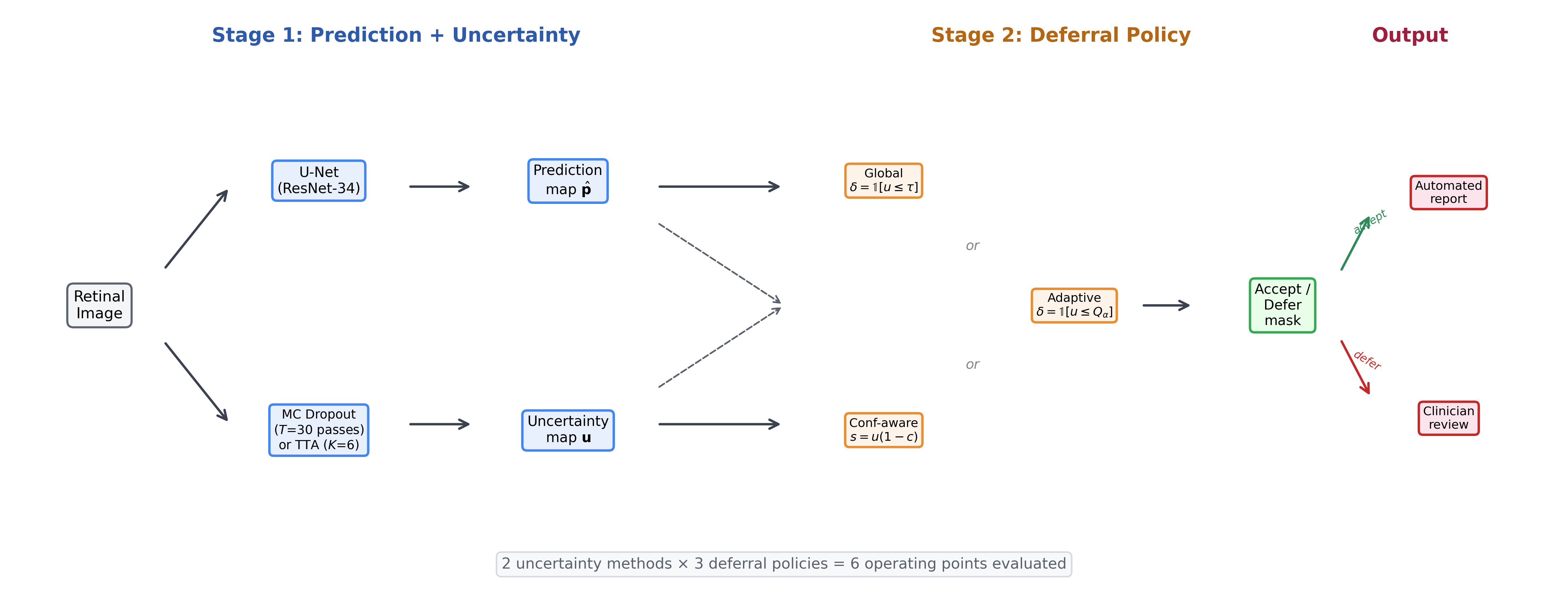}
\caption{Overview of the two-stage decision pipeline. Stage 1 produces a
segmentation prediction and uncertainty estimate. Stage 2 applies a deferral
policy to produce an accept/defer decision per pixel. The two stages are
modular: any uncertainty method can pair with any deferral policy.}
\label{fig:pipeline}
\end{figure}

Figure~\ref{fig:pipeline} illustrates the full pipeline; read it as an
estimation block feeding an independently swappable decision block. We
formalize it in four parts: the problem setting (Section~\ref{sec:formulation}),
uncertainty estimation (Section~\ref{sec:uncertainty}), the deferral framework
(Section~\ref{sec:deferral}), and calibration (Section~\ref{sec:calibration}).

\subsection{Problem formulation}
\label{sec:formulation}

Let $\mathbf{x} \in \mathbb{R}^{H \times W \times C}$ be an input image and
$\mathbf{y} \in \{0, 1\}^{H \times W}$ the corresponding binary segmentation
mask (vessel vs.\ background). A segmentation model $f_\theta$ produces a
probability map $\hat{\mathbf{p}} = f_\theta(\mathbf{x}) \in [0, 1]^{H \times
W}$, where $\hat{p}_{ij}$ is the predicted probability that pixel $(i, j)$ is
a vessel. The hard prediction is $\hat{y}_{ij} = \mathbb{1}[\hat{p}_{ij} >
0.5]$.

An uncertainty estimation procedure augments this pipeline with an uncertainty
map $\mathbf{u} \in \mathbb{R}_{\geq 0}^{H \times W}$. We use $u_{ij}$ to
denote the uncertainty assigned to pixel $(i, j)$. Higher values of $u_{ij}$
should, ideally, correspond to higher probability of error, i.e.,
$\hat{y}_{ij} \neq y_{ij}$.

A \emph{deferral policy} $\delta: \mathbb{R}_{\geq 0}^{H \times W} \times
[0,1]^{H \times W} \to \{0, 1\}^{H \times W}$ maps the uncertainty and
prediction to a binary decision map, where $\delta_{ij} = 1$ means ``accept
the prediction at pixel $(i,j)$'' and $\delta_{ij} = 0$ means ``defer this
pixel to a human reviewer.'' The goal is a policy that accepts correct
predictions and defers incorrect ones, subject to a coverage constraint: the
fraction of accepted pixels, $\text{coverage} = \frac{1}{HW}\sum_{ij}
\delta_{ij}$, should remain high enough that deferral is practical.

We evaluate deferral quality by the \emph{error reduction ratio}:
\begin{equation}
  \text{ERR} = \frac{e_{\text{before}} - e_{\text{after}}}{e_{\text{before}}}
  \label{eq:err}
\end{equation}
where $e_{\text{before}}$ is the error rate on all pixels and
$e_{\text{after}}$ is the error rate on accepted pixels only. A good deferral
policy achieves high ERR at high coverage.

\subsection{Uncertainty estimation}
\label{sec:uncertainty}

\subsubsection{MC Dropout}

MC Dropout \citep{gal2016dropout} keeps dropout active at test time and
performs $T$ stochastic forward passes through the network. For input
$\mathbf{x}$, let $\hat{p}_{ij}^{(t)} = f_{\theta_t}(\mathbf{x})_{ij}$ be
the predicted probability at pixel $(i,j)$ on pass $t$, where $\theta_t$
denotes the weights with dropout mask $t$ applied. The mean prediction is
\begin{equation}
  \bar{p}_{ij} = \frac{1}{T} \sum_{t=1}^{T} \hat{p}_{ij}^{(t)}.
  \label{eq:mc_mean}
\end{equation}

We quantify uncertainty using mutual information (MI), which decomposes total
uncertainty into aleatoric and epistemic components. For a binary prediction:
\begin{equation}
  \text{MI}_{ij} = H(\bar{p}_{ij}) - \frac{1}{T} \sum_{t=1}^{T} H(\hat{p}_{ij}^{(t)})
  \label{eq:mi}
\end{equation}
where $H(p) = -p \log p - (1-p) \log(1-p)$ is the binary entropy. The first
term is the predictive entropy (total uncertainty); the second is the expected
entropy across passes (aleatoric uncertainty). Their difference, MI, isolates
epistemic uncertainty: disagreement among the stochastic passes about the
correct class.

MI is preferable to raw variance for binary outputs because variance is bounded
by $0.25$ and is a symmetric, single-peaked function of $\bar{p}$, which
conflates high uncertainty with predictions near $p = 0.5$ regardless of
whether the model's passes agree. MI, by contrast, is zero when all passes
agree (even if they all predict $p = 0.5$) and large when passes disagree.

We use $T = 30$ passes with dropout probability $p_{\text{drop}} = 0.3$,
applied as spatial dropout (Dropout2d) after the decoder and before the
segmentation head. Inference is batched in chunks of 5 passes to manage GPU
memory.

\subsubsection{Test-Time Augmentation}

TTA applies a set of deterministic geometric transformations
$\{\tau_k\}_{k=1}^{K}$ to the input, performs a forward pass on each
transformed input, applies the inverse transformation $\tau_k^{-1}$ to the
output, and aggregates:
\begin{equation}
  \bar{p}_{ij}^{\text{TTA}} = \frac{1}{K} \sum_{k=1}^{K} \left[\tau_k^{-1}
  \circ f_\theta \circ \tau_k\right](\mathbf{x})_{ij}.
  \label{eq:tta_mean}
\end{equation}

We use $K = 6$ transformations: identity, horizontal flip, vertical flip,
90$^\circ$ rotation, 180$^\circ$ rotation, and 270$^\circ$ rotation. These
are exact (no interpolation artifacts) and cover the symmetry group of
the square.

Uncertainty is the pixel-wise variance across augmented predictions:
\begin{equation}
  u_{ij}^{\text{TTA}} = \frac{1}{K} \sum_{k=1}^{K} \left(\hat{p}_{ij}^{(k)} -
  \bar{p}_{ij}^{\text{TTA}}\right)^2
  \label{eq:tta_var}
\end{equation}
where $\hat{p}_{ij}^{(k)} = [\tau_k^{-1} \circ f_\theta \circ
\tau_k](\mathbf{x})_{ij}$.

We also compute the entropy of the mean prediction as an additional uncertainty
signal:
\begin{equation}
  u_{ij}^{\text{TTA-ent}} = H(\bar{p}_{ij}^{\text{TTA}}).
  \label{eq:tta_entropy}
\end{equation}

The two methods probe different sources of instability. MC Dropout samples
weight perturbations (epistemic uncertainty); TTA tests whether predictions
are stable under input transforms that should not change the answer. A
vessel is a vessel regardless of image orientation. Section~\ref{sec:discussion}
discusses why this distinction matters for deferral quality.

\subsection{Deferral framework}
\label{sec:deferral}

Given an uncertainty map $\mathbf{u}$ and a prediction map $\hat{\mathbf{p}}$,
we define three deferral policies.

\subsubsection{Global threshold deferral}

The simplest policy defers all pixels whose uncertainty exceeds a fixed
threshold $\tau$:
\begin{equation}
  \delta_{ij}^{\text{global}} = \mathbb{1}[u_{ij} \leq \tau].
  \label{eq:global}
\end{equation}

The threshold $\tau$ is selected on a validation set by sweeping over
percentiles of the uncertainty distribution and choosing the value that
maximizes a deferral-specific F1 score:
\begin{equation}
  F1_{\text{def}} = \frac{2 \cdot \text{prec}_{\text{def}} \cdot
  \text{rec}_{\text{def}}}{\text{prec}_{\text{def}} +
  \text{rec}_{\text{def}}}
  \label{eq:def_f1}
\end{equation}
where $\text{prec}_{\text{def}}$ is the fraction of deferred pixels that are
actual errors, and $\text{rec}_{\text{def}}$ is the fraction of all error
pixels that are deferred.

Global thresholding has the advantage of simplicity and a single
interpretable parameter. Its weakness is that a single $\tau$ must work
across images of varying difficulty: an easy image has low uncertainty
everywhere, so few pixels are deferred, while a hard image may have most
of its uncertainty mass above $\tau$, causing excessive deferral.

\subsubsection{Image-adaptive threshold deferral}

To address the variable-difficulty problem, image-adaptive thresholding
computes a separate threshold for each image based on its own uncertainty
distribution:
\begin{equation}
  \tau_n = Q_\alpha(\{u_{ij} : (i,j) \in \text{image } n\})
  \label{eq:adaptive_threshold}
\end{equation}
where $Q_\alpha$ is the $\alpha$-th percentile. The percentile $\alpha$
is chosen on a validation set; at test time, each image's threshold is
its own $\alpha$-th uncertainty percentile.

The deferral rule is then:
\begin{equation}
  \delta_{ij}^{\text{adaptive}} = \mathbb{1}[u_{ij} \leq \tau_n]
  \quad \text{for pixel } (i,j) \text{ in image } n.
  \label{eq:adaptive}
\end{equation}

This guarantees a fixed fraction $\alpha / 100$ of pixels are deferred
per image, which may be desirable in clinical workflows with a fixed
review budget. The trade-off is that easy images have pixels deferred
unnecessarily (the bottom $1-\alpha$ fraction of uncertainty in an
easy image may all be correct), while hard images may still not defer
enough.

We fit $\alpha$ using one of two criteria on the validation set:
\begin{itemize}[leftmargin=*]
  \item \emph{Max-F1}: choose $\alpha$ to maximize $F1_{\text{def}}$.
  \item \emph{Coverage-Dice}: choose the highest coverage (lowest deferral
  rate) such that Dice on accepted pixels exceeds a clinical threshold
  (0.82 in our experiments).
\end{itemize}

\subsubsection{Confidence-aware deferral}

Global and adaptive thresholding use only the uncertainty map $\mathbf{u}$.
But the prediction map $\hat{\mathbf{p}}$ contains additional information:
a pixel with $\hat{p}_{ij} = 0.51$ is near the decision boundary and likely
to flip under small perturbations, while $\hat{p}_{ij} = 0.95$ is far from
the boundary even if its uncertainty is moderate.

We define a prediction confidence score:
\begin{equation}
  c_{ij} = 2 \left| \hat{p}_{ij} - 0.5 \right| \in [0, 1]
  \label{eq:confidence}
\end{equation}
which is 0 when the prediction is maximally ambiguous ($\hat{p} = 0.5$) and
1 when the model is fully committed ($\hat{p} \in \{0, 1\}$).

The confidence-aware deferral score combines uncertainty and confidence:
\begin{equation}
  s_{ij} = u_{ij} \cdot (1 - c_{ij})
  \label{eq:conf_score}
\end{equation}

Pixels with high uncertainty \emph{and} low confidence receive the highest
scores and are deferred first. Pixels that are uncertain but confident
(the model predicts $p = 0.95$ but with high variance across passes)
receive lower scores and are retained.

The deferral rule is:
\begin{equation}
  \delta_{ij}^{\text{conf}} = \mathbb{1}[s_{ij} \leq \tau_s]
  \label{eq:conf_deferral}
\end{equation}
where $\tau_s$ is again chosen on a validation set.

A concrete example clarifies this. Consider two pixels, both with
$u_{ij} = 0.05$. Pixel A has $\hat{p} = 0.52$ (barely classified as
vessel), so $c_A = 0.04$ and $s_A = 0.05 \times 0.96 = 0.048$. Pixel B
has $\hat{p} = 0.92$ (clearly vessel), so $c_B = 0.84$ and
$s_B = 0.05 \times 0.16 = 0.008$. Global thresholding treats them
identically; confidence-aware deferral correctly prioritizes pixel A
for deferral.

\subsection{Calibration via temperature scaling}
\label{sec:calibration}

Temperature scaling \citep{guo2017calibration} learns a single parameter
$T > 0$ on a validation set to rescale the model's logits:
\begin{equation}
  \hat{p}_{ij}^{\text{cal}} = \sigma\left(\frac{\text{logit}(\hat{p}_{ij})}{T}\right)
  \label{eq:temp_scaling}
\end{equation}
where $\sigma$ is the sigmoid function and $\text{logit}(p) = \log(p / (1-p))$.

When $T > 1$, the sigmoid is flattened and predictions move toward 0.5
(the model becomes less confident). When $T < 1$, predictions sharpen.
$T$ is optimized by minimizing BCE loss on validation logits:
\begin{equation}
  T^* = \arg\min_T \; -\frac{1}{N_{\text{val}}} \sum_{ij}
  \left[ y_{ij} \log \hat{p}_{ij}^{\text{cal}} +
  (1 - y_{ij}) \log (1 - \hat{p}_{ij}^{\text{cal}}) \right]
  \label{eq:temp_opt}
\end{equation}
using L-BFGS, with $T$ constrained to $[0.05, +\infty)$.

In our experiments, the learned temperatures are $T = 1.19$ (MC Dropout) and
$T = 1.35$ (TTA), both $> 1$, confirming mild overconfidence. ECE changes
are negligible (Section~\ref{sec:results_calibration}), and the effect on
deferral quality is neutral to negative.

Why would calibration fail to help deferral? Temperature scaling is a
monotonic transformation of predicted probabilities: it rescales magnitudes
without changing rankings. If pixel A has higher uncertainty than pixel B
before calibration, it still does after. Deferral policies that threshold on
uncertainty ranking are therefore invariant to temperature scaling. The only
indirect path is through the confidence score $c_{ij}$ in the
confidence-aware policy, since softening predictions shifts which pixels sit
near the decision boundary. In practice, this second-order effect is small.
Section~\ref{sec:results_calibration} presents the evidence.

\section{Experimental setup}
\label{sec:experimental_setup}

\subsection{Datasets}

\paragraph{DRIVE.} The Digital Retinal Images for Vessel Extraction dataset
\citep{staal2004drive} contains 40 color fundus photographs (565 $\times$ 584
pixels) with manually annotated binary vessel masks. We use the standard
20/20 train/test split. From the 20 training images, we hold out 4 for
validation (used for threshold selection and temperature fitting).

\paragraph{STARE.} The Structured Analysis of the Retina dataset
\citep{hoover2000stare} contains 20 fundus images (700 $\times$ 605 pixels)
with vessel annotations by two experts. We use the first expert's annotations
and evaluate on all 20 images in a zero-shot cross-dataset setting (no
fine-tuning on STARE).

\paragraph{CHASE\_DB1.} The Child Heart and Health Study in England dataset
\citep{fraz2012chase} contains 28 fundus images (999 $\times$ 960 pixels)
from pediatric subjects, with vessel annotations. We again evaluate in a
zero-shot setting using all 28 images.

Using STARE and CHASE\_DB1 as external test sets tests whether uncertainty
quality, not just segmentation accuracy, transfers under domain shift. The
three datasets differ in resolution, patient population (adults vs.\ children),
imaging equipment, and annotation protocols.

\subsection{Preprocessing and augmentation}

All images are resized to 512 $\times$ 512 pixels. During training, we
extract 256 $\times$ 256 patches with vessel-aware sampling: 80\% of patches
are constrained to contain at least 16 vessel pixels, addressing the heavy
class imbalance in retinal images (vessels typically occupy 10--15\% of pixels).
We extract 32 patches per image per epoch. Training augmentations include
random horizontal and vertical flips, rotations up to 15$^\circ$, and
brightness/contrast jittering.

At test time, we process full 512 $\times$ 512 images (no patching) to
avoid boundary artifacts in the uncertainty maps.

\subsection{Model and training}

The base model is a U-Net \citep{ronneberger2015unet} with a ResNet-34
encoder \citep{he2016deep}, pretrained on ImageNet, implemented using
the \texttt{segmentation-models-pytorch} library. Dropout2d with
$p = 0.3$ is applied after the decoder and before the final $1 \times 1$
convolution.

Training uses the Adam optimizer \citep{kingma2015adam} with learning rate
$10^{-4}$, weight decay $10^{-4}$, and gradient clipping at norm 1.0.
The loss function is a hybrid of Dice loss and binary cross-entropy:
\begin{equation}
  \mathcal{L} = \frac{1}{2}\mathcal{L}_{\text{Dice}} +
  \frac{1}{2}\mathcal{L}_{\text{BCE}}
  \label{eq:loss}
\end{equation}
with a positive class weight of 4.0 in the BCE term to compensate for
class imbalance. The Dice loss uses a smoothing constant of 1.0.

We train for 80 epochs with batch size 6. The model converges around epoch
60; we select the checkpoint with the best validation Dice.

For the ensemble baseline, we train 5 U-Net models with identical
hyperparameters but different random seeds (controlling weight
initialization, data shuffling, and dropout masks).

\subsection{Evaluation protocol}

All metrics are computed on the DRIVE test set (20 images) unless otherwise
stated. For each image, we compute the mean prediction and uncertainty map
using the specified method (MC Dropout with $T = 30$ passes, or TTA with
$K = 6$ augmentations). Deferral thresholds are fit on the validation set
and applied without modification to the test set.

We report results for six configurations, covering all combinations of
\{MC Dropout, TTA\} $\times$ \{global, adaptive, confidence-aware\} deferral,
plus additional experiments with temperature scaling.

\subsection{Metrics}

\paragraph{Segmentation quality.}
\begin{align}
  \text{Dice} &= \frac{2 \sum_{ij} \hat{y}_{ij} y_{ij}}
  {2\sum_{ij} \hat{y}_{ij} y_{ij} + \sum_{ij} \hat{y}_{ij}(1 - y_{ij}) +
  \sum_{ij} (1 - \hat{y}_{ij}) y_{ij}}
  \label{eq:dice} \\
  \text{IoU} &= \frac{\sum_{ij} \hat{y}_{ij} y_{ij}}
  {\sum_{ij} \hat{y}_{ij} y_{ij} + \sum_{ij} \hat{y}_{ij}(1 - y_{ij}) +
  \sum_{ij} (1 - \hat{y}_{ij}) y_{ij}}
  \label{eq:iou} \\
  \text{AUC} &= \text{Area under the ROC curve, computed on } \hat{p}_{ij}
  \text{ vs.\ } y_{ij}
  \label{eq:auc}
\end{align}

\paragraph{Calibration.}
\begin{equation}
  \text{ECE} = \sum_{b=1}^{B} \frac{n_b}{N}
  \left| \text{acc}(b) - \text{conf}(b) \right|
  \label{eq:ece}
\end{equation}
where $B = 15$ bins, $n_b$ is the number of pixels in bin $b$,
$\text{acc}(b)$ is the fraction of correctly predicted pixels in bin $b$,
and $\text{conf}(b)$ is the mean predicted probability in bin $b$.

\paragraph{Uncertainty quality.}
\begin{equation}
  \text{Unc-AUROC} = \text{AUC}(u_{ij}, \; \mathbb{1}[\hat{y}_{ij} \neq y_{ij}])
  \label{eq:unc_auroc}
\end{equation}
This is the ROC AUC when uncertainty is used as a binary classifier of
prediction error. An Unc-AUROC of 0.5 means uncertainty is no better than
random at identifying errors; 1.0 means it perfectly separates correct from
incorrect pixels.

\paragraph{Risk-coverage metrics.} For a given deferral threshold, we compute
the Dice score on accepted pixels only and plot it against coverage (fraction
of accepted pixels). The Area Under the Coverage Curve (AUCC) summarizes this
trade-off in a single number:
\begin{equation}
  \text{AUCC} = \int_0^1 \text{Dice}(\text{coverage}) \; d(\text{coverage})
  \label{eq:aucc}
\end{equation}
Higher AUCC means the method's uncertainty is more effective at identifying
which pixels to keep.

\paragraph{Deferral metrics.} Coverage and error reduction ratio (ERR) are
defined in Section~\ref{sec:formulation} (Equation~\ref{eq:err}).

\section{Results}
\label{sec:results}

\begin{figure}[H]
\centering
\includegraphics[width=0.92\textwidth]{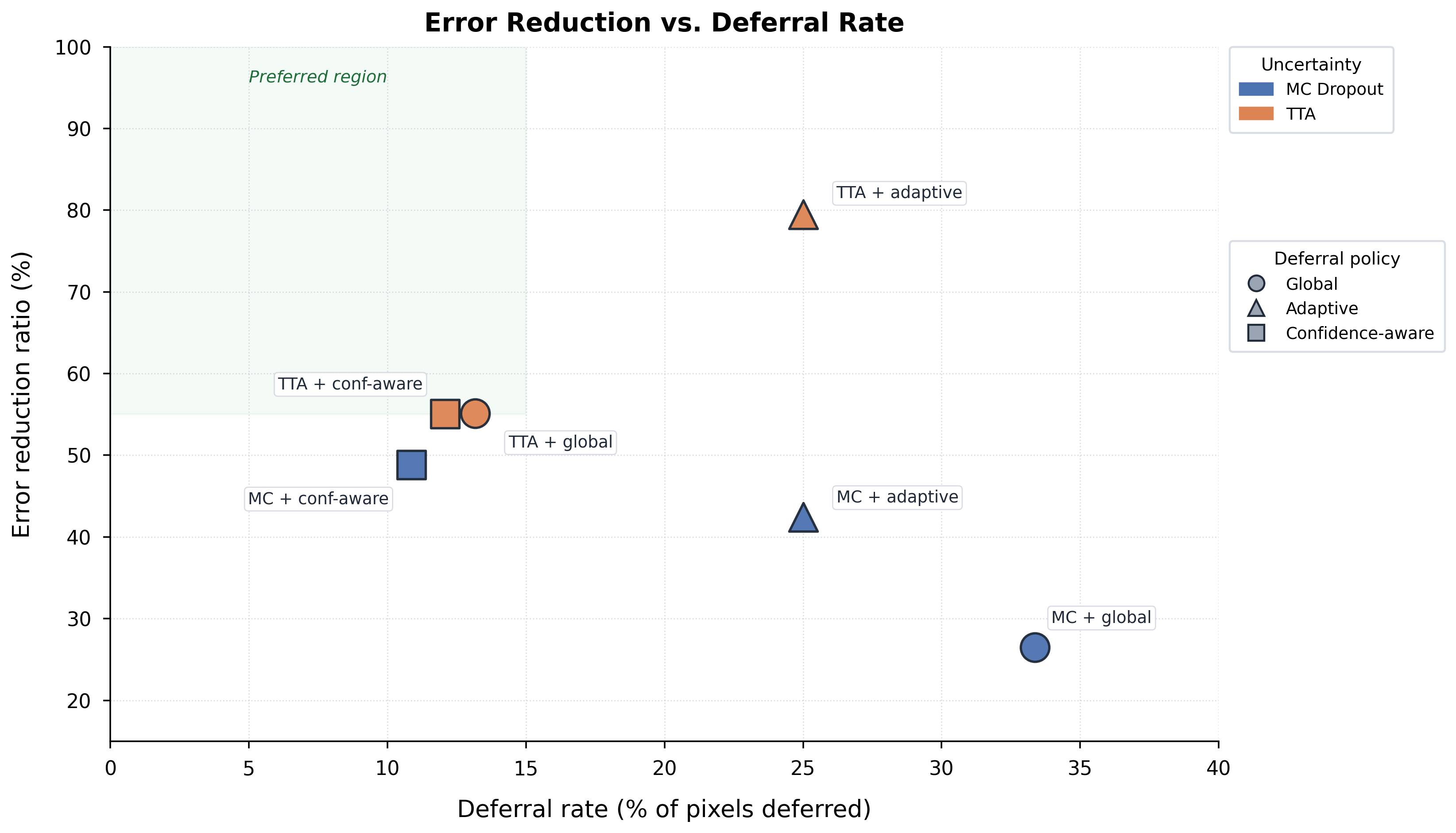}
\caption{Error reduction vs.\ deferral rate. Each point pairs an uncertainty
method (MC Dropout in blue, TTA in orange) with a deferral policy (marker
shape). TTA + adaptive dominates the upper-left (79.5\% error reduction at
25\% deferral); TTA + confidence-aware is the Pareto optimum at low budgets
(55\% at only 12\%). Every TTA configuration outperforms every MC Dropout
configuration---the decision layer reshapes, not merely refines, what the
uncertainty estimator delivers.}
\label{fig:summary_scatter}
\end{figure}

We organize results around a single question: \emph{how much does the
decision layer matter?} Five claims answer it. The headline: TTA paired
with adaptive deferral removes roughly 80\% of segmentation errors at a
25\% deferral rate (Table~\ref{tab:main_results}); the same uncertainty
under a naive global rule removes only a third of that.
Figure~\ref{fig:summary_scatter} plots every method--policy combination
on one axis and makes the interaction visible at a glance---the spread
across configurations is far wider than the spread across uncertainty
sources alone, which establishes the decision layer as the dominant
contributor to final decision quality.

\begin{table}[H]
\centering
\caption{Main results on the DRIVE test set (20 images). TTA matches the
ensemble on segmentation quality while producing the best uncertainty
(Unc-AUROC 0.881) and the largest error reduction (79.5\% at 25\% deferral)
at 3.1$\times$ the speed of MC Dropout. Best values per column in bold;
runtime is per image on a single GPU. ERR $=$ error reduction ratio;
Def.\ \% $=$ fraction of pixels deferred.}
\label{tab:main_results}
\scriptsize
\setlength{\tabcolsep}{3.5pt}
\renewcommand{\arraystretch}{1.03}
\resizebox{\textwidth}{!}{%
\begin{tabular}{l c c c c c c c c}
\toprule
\textbf{Method} & \textbf{Dice} & \textbf{AUC} & \textbf{IoU} &
\textbf{ECE} & \textbf{Unc-AUROC} & \textbf{ERR} &
\textbf{Def.\ \%} & \textbf{Time (s)} \\
\midrule
Deterministic            & 0.736 & 0.944 & ---   & 0.245 & 0.500 & ---           & ---   & \textbf{0.08} \\
MC Drop.\ (global)       & 0.763 & 0.965 & 0.617 & 0.035 & 0.722 & 26.5\%        & 33.4\% & 2.65 \\
MC Drop.\ (adaptive)     & 0.763 & 0.965 & 0.617 & 0.035 & 0.722 & 42.4\%        & 25.0\% & 2.65 \\
MC Drop.\ (conf-aware)   & 0.763 & 0.965 & 0.617 & 0.035 & 0.722 & 48.8\%        & 10.9\% & 2.65 \\
\addlinespace[2pt]
TTA (global)             & 0.768 & 0.969 & 0.624 & 0.035 & \textbf{0.881} & 55.1\%        & 13.2\% & 0.86 \\
TTA (adaptive)           & \textbf{0.768} & \textbf{0.969} & \textbf{0.624} & 0.035 & \textbf{0.881} & \textbf{79.5\%} & 25.0\% & 0.86 \\
TTA (conf-aware)         & 0.768 & 0.969 & 0.624 & 0.035 & \textbf{0.881} & 55.1\%        & \textbf{12.1\%} & 0.86 \\
\midrule
Ensemble ($N{=}5$)       & 0.771 & 0.947 & 0.627 & \textbf{0.031} & 0.833 & ---           & ---   & 3.90 \\
\bottomrule
\end{tabular}%
}
\end{table}

\paragraph{Reading the table.} Rows 2--4 and 5--7 share identical Dice,
AUC, and Unc-AUROC within each uncertainty method; only the rightmost
columns (ERR, Def.\,\%) change. The three-fold spread in error reduction
within a single uncertainty source is the clearest evidence that the
decision layer is not a formality---it is where most of the decision
quality is made or lost.

\subsection{TTA produces superior uncertainty for deferral}
\label{sec:results_uncertainty}

\textbf{Claim:} TTA uncertainty discriminates correct from incorrect
predictions substantially better than MC Dropout, while requiring less
computation.

\emph{Insight before the numbers.} If errors concentrate at specific
geometric structures, an uncertainty source that probes geometric
instability should separate them more cleanly than one that probes
weight instability.

TTA achieves an Unc-AUROC of 0.881, compared to 0.722 for MC Dropout
(Table~\ref{tab:main_results}). This is a large gap: TTA correctly ranks
88.1\% of error/non-error pixel pairs by uncertainty, versus 72.2\% for
MC Dropout. The difference is consistent across images and robust to
bootstrap resampling.

TTA also produces higher AUCC for both Dice (0.846 vs.\ 0.792) and AUC
(0.888 vs.\ 0.884), meaning that progressively removing the most uncertain
pixels improves performance faster with TTA uncertainty.

In terms of computation, TTA requires 6 forward passes with deterministic
augmentations ($K = 6$), while MC Dropout requires 30 stochastic passes
($T = 30$). TTA inference takes approximately 0.86\,s per image versus
2.65\,s for MC Dropout---a 3.1$\times$ speedup.

\paragraph{Interpretation.} The quality gap has a structural explanation.
In retinal vessel segmentation, errors concentrate at vessel boundaries
and thin vessel tips---regions sensitive to small input perturbations. TTA
directly tests this sensitivity by applying geometric transforms; if a
boundary pixel flips under rotation, TTA flags it. MC Dropout instead
samples weight perturbations, producing more spatially diffuse uncertainty
that does not concentrate as sharply on error-prone structures. The
takeaway generalizes beyond retinas: decision-time uncertainty should be
sourced from perturbations that mimic the error mode, not from whichever
Bayesian approximation is easiest to fit. ``Better uncertainty'' is not a
fixed quantity---it is the uncertainty whose failure mode matches the
decision rule downstream.

\begin{figure}[H]
\centering
\includegraphics[width=\textwidth]{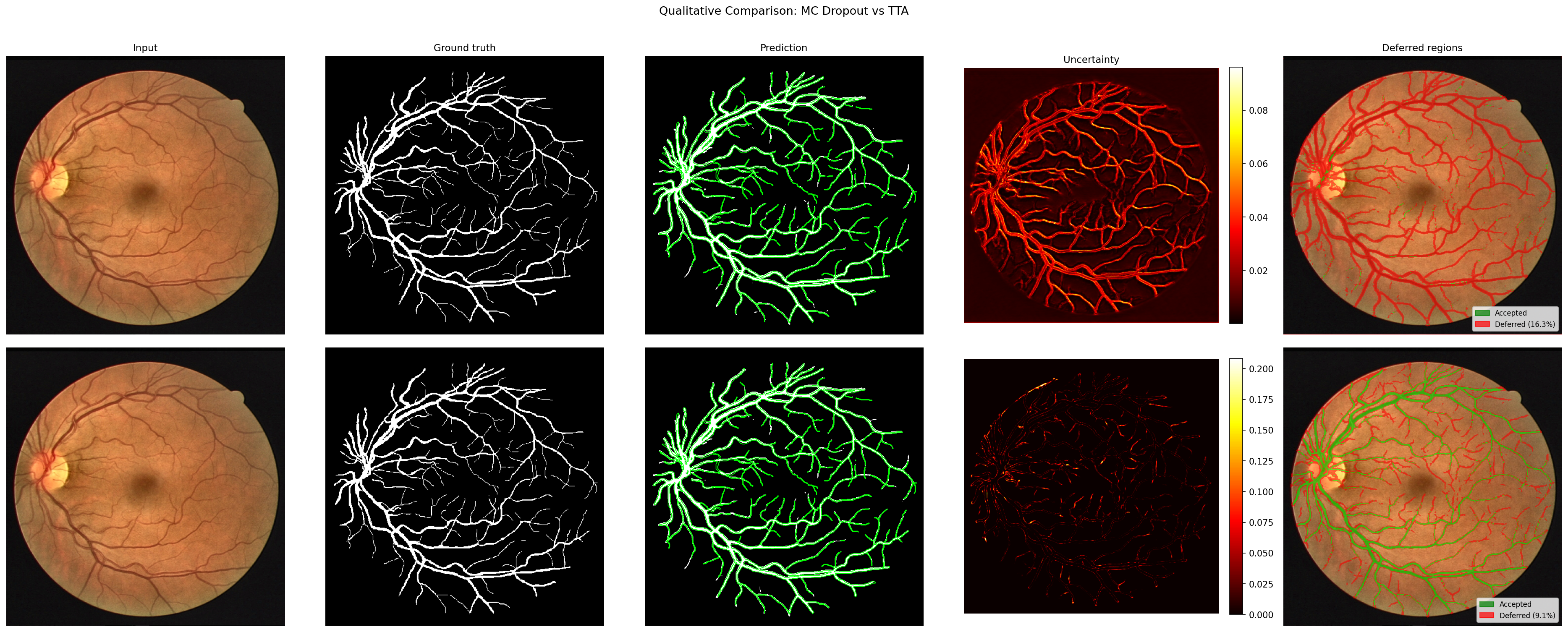}
\caption{Uncertainty maps from MC Dropout and TTA. TTA uncertainty
concentrates on boundary and thin-vessel regions where errors are most
frequent, while MC Dropout produces more spatially diffuse uncertainty.}
\label{fig:uncertainty_maps}
\end{figure}

\subsection{Adaptive deferral maximizes error removal; confidence-aware deferral is most efficient at low budget}
\label{sec:results_deferral}

\textbf{Claim:} Adaptive deferral achieves the largest absolute error removal,
while confidence-aware deferral is the most efficient policy at low review
budgets.

Table~\ref{tab:deferral_comparison} breaks down the deferral results. For MC
Dropout, global thresholding defers 33.4\% of pixels and reduces error by
26.5\%. Adaptive deferral improves to 42.4\% error reduction at a 25\% deferral
budget, and confidence-aware deferral reaches 48.8\% while deferring only
10.9\% of pixels.

For TTA, adaptive deferral yields the headline result: 79.5\% error reduction
at 25.0\% deferral. The confidence-aware policy matches the global TTA policy
in error reduction but with a smaller review budget (12.1\% vs.\ 13.2\%
deferred).

\begin{table}[H]
\centering
\caption{Deferral strategy comparison on the DRIVE test set. Confidence-aware
deferral dominates at low review budgets---it delivers 4.56 percentage points
of error reduction per percentage point deferred, roughly $5.8\times$ the
efficiency of the MC Dropout global baseline. Adaptive deferral produces the
largest absolute error removal (79.5\% at 25\% deferral). Bold marks the best
value in each column within its uncertainty group.}
\label{tab:deferral_comparison}
\scriptsize
\setlength{\tabcolsep}{3.5pt}
\renewcommand{\arraystretch}{1.03}
\begin{center}
\begin{tabular}{l l c c c c c}
\toprule
\textbf{Uncertainty} & \textbf{Deferral} & \textbf{Error before} &
\textbf{Error after} & \textbf{ERR} & \textbf{Def.\ \%} &
\textbf{ERR / Def.\ \%} \\
\midrule
MC Dropout & Global      & 0.0658 & 0.0484 & 26.5\% & 33.4\% & 0.79 \\
MC Dropout & Adaptive    & 0.0658 & 0.0379 & 42.4\% & 25.0\% & 1.70 \\
MC Dropout & Conf-aware  & 0.0658 & \textbf{0.0337} & \textbf{48.8\%} & \textbf{10.9\%} & \textbf{4.49} \\
\addlinespace[2pt]
\midrule
\addlinespace[2pt]
TTA        & Global      & 0.0640 & 0.0287 & 55.1\% & 13.2\% & 4.19 \\
TTA        & Adaptive    & 0.0640 & \textbf{0.0131} & \textbf{79.5\%} & 25.0\% & 3.18 \\
TTA        & Conf-aware  & 0.0640 & 0.0288 & 55.1\% & \textbf{12.1\%} & \textbf{4.56} \\
\bottomrule
\end{tabular}
\end{center}
\end{table}

The rightmost column in Table~\ref{tab:deferral_comparison}, ERR per
percentage point deferred (ERR/Def.\ \%), quantifies deferral efficiency.
For MC Dropout, confidence-aware deferral is dramatically more efficient than
the global policy (4.49 vs.\ 0.79). For TTA, both global and confidence-aware
policies are efficient, with confidence-aware slightly better (4.56 vs.\ 4.19)
at a lower deferral rate.

\paragraph{Interpretation.} Global thresholding wastes deferral budget on
pixels that are uncertain but correctly predicted. The confidence-aware
score $s = u \cdot (1 - c)$ suppresses these false alarms. Vessel
interiors are a clear example: MC Dropout assigns moderate uncertainty
there, but predictions sit well above 0.5, so $c$ is high and $s$ remains
low. The broader lesson: deferral efficiency is decided in the score
function, not just in the uncertainty source. A decision rule that treats
``uncertain'' and ``borderline'' as synonyms is leaving free error
reduction on the table.

\begin{figure}[H]
\centering
\includegraphics[width=\textwidth]{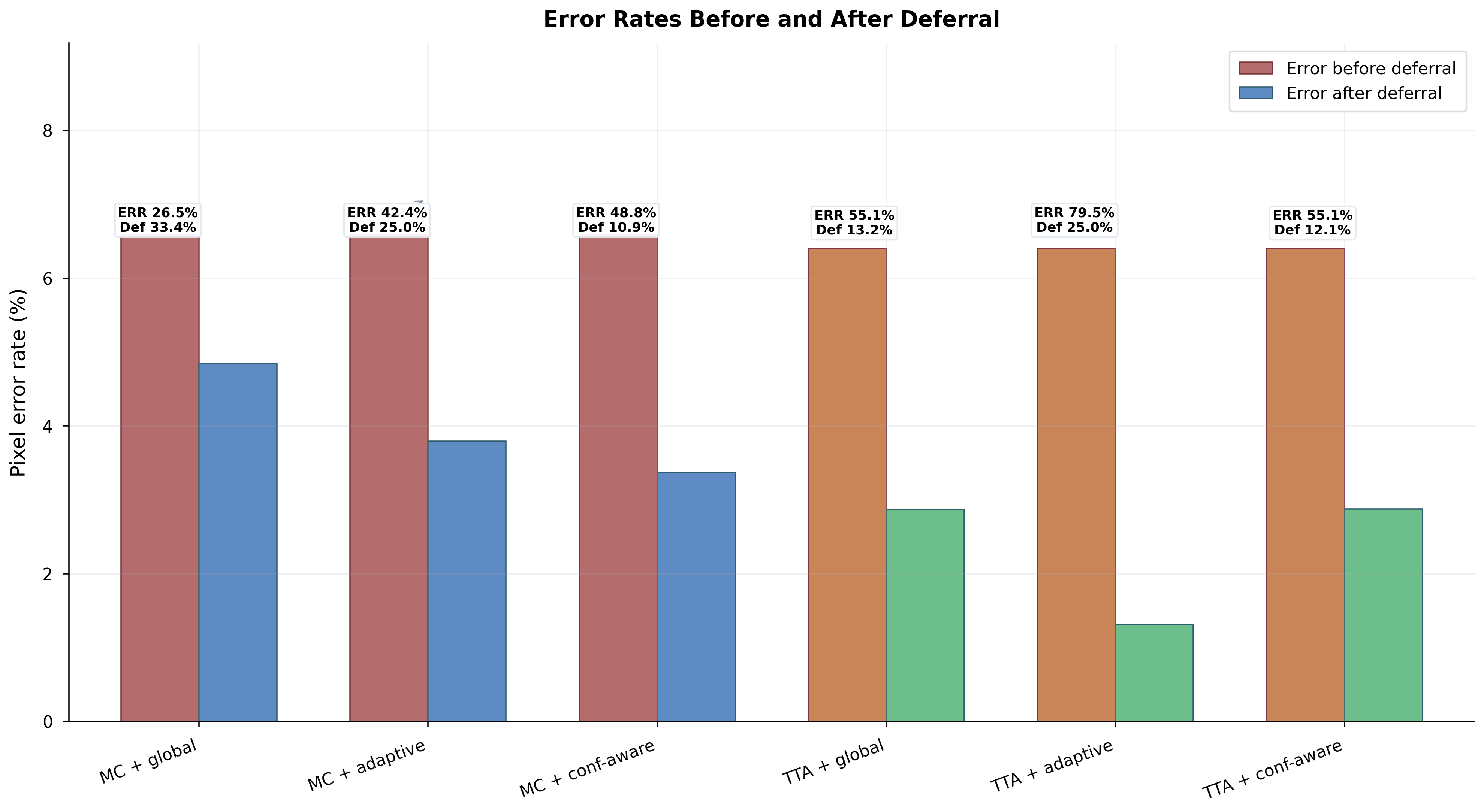}
\caption{Error rates before and after deferral across all configurations.
TTA+adaptive achieves 80\% error reduction; TTA+conf-aware achieves 55\%
error reduction at roughly half the review budget. The choice of both
uncertainty method and deferral policy substantially reshapes the practical
review workflow.}
\label{fig:error_reduction_bars}
\end{figure}

\begin{figure}[H]
\centering
\begin{subfigure}[b]{0.49\textwidth}
  \centering
  \includegraphics[width=\linewidth]{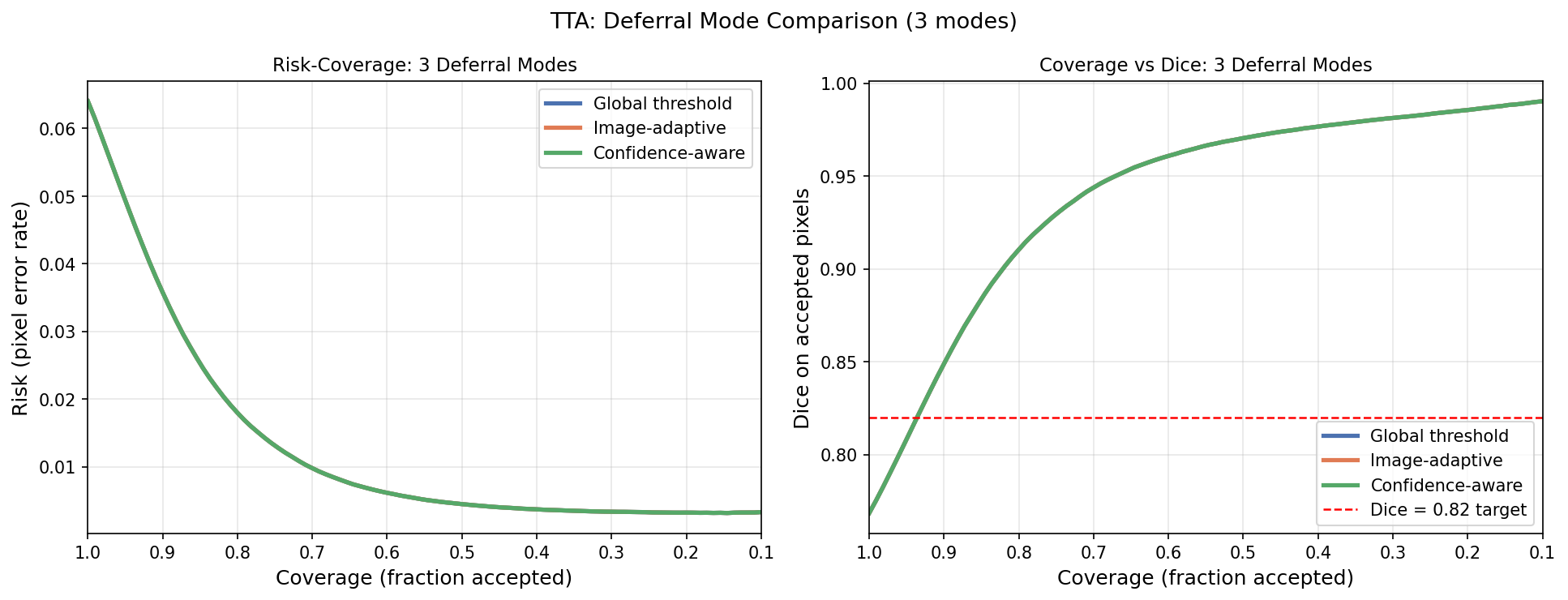}
  \caption{TTA risk-coverage under three deferral modes. All three policies
  reduce risk monotonically; confidence-aware reaches the clinical Dice
  target (0.82) at the highest coverage.}
  \label{fig:deferral_curves}
\end{subfigure}\hfill
\begin{subfigure}[b]{0.49\textwidth}
  \centering
  \includegraphics[width=\linewidth]{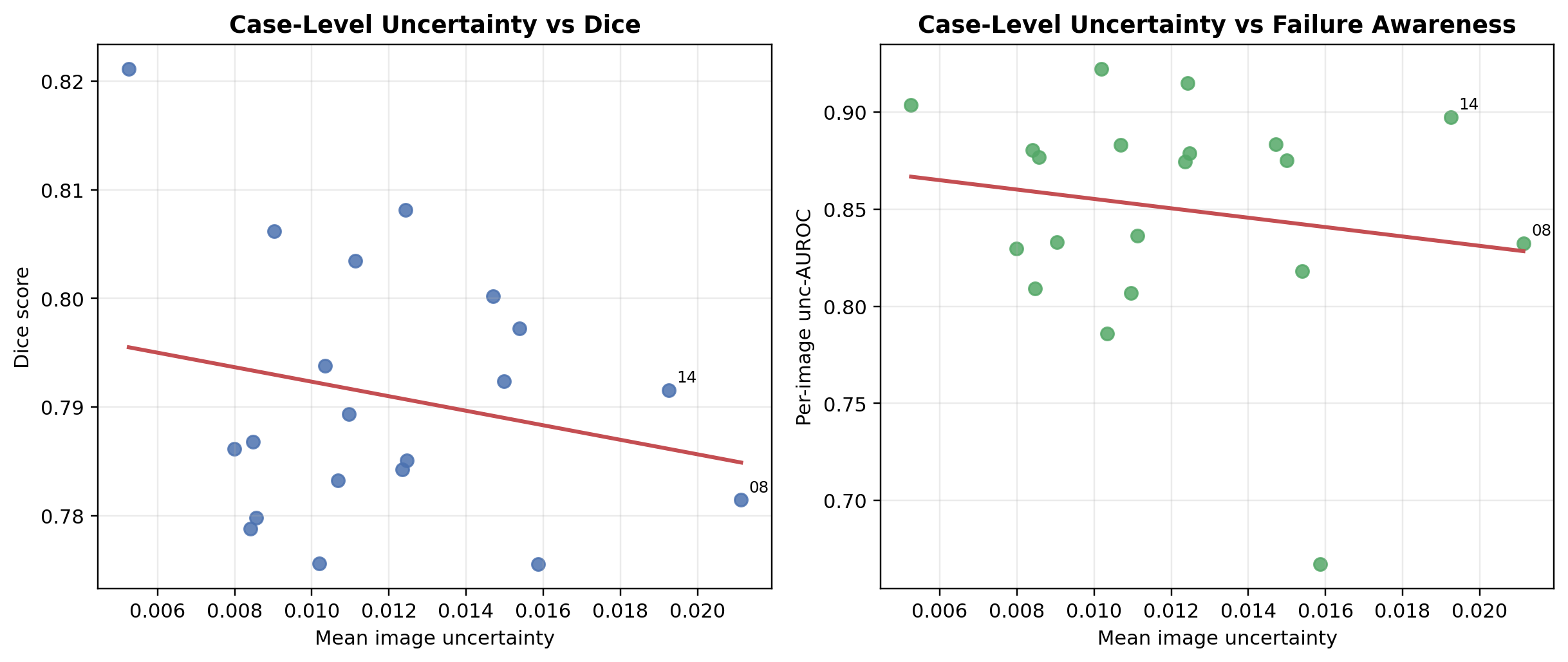}
  \caption{Case-level analysis. Left: higher mean uncertainty correlates
  with lower Dice. Right: per-image Unc-AUROC remains high even for
  difficult cases---uncertainty is most informative where it matters most.}
  \label{fig:unc_vs_error}
\end{subfigure}
\caption{Deferral behavior and case-level uncertainty analysis. Left panel
shows TTA deferral modes; right panel confirms that uncertainty quality
holds across the difficulty spectrum.}
\label{fig:deferral_views}
\end{figure}

\subsection{Risk-coverage analysis}
\label{sec:results_risk_coverage}

\textbf{Claim:} Selective prediction using TTA uncertainty produces a
steeper risk-coverage curve than MC Dropout and achieves clinically
acceptable Dice ($\geq 0.82$) at 90\% coverage.

\begin{figure}[H]
\centering
\includegraphics[width=\textwidth]{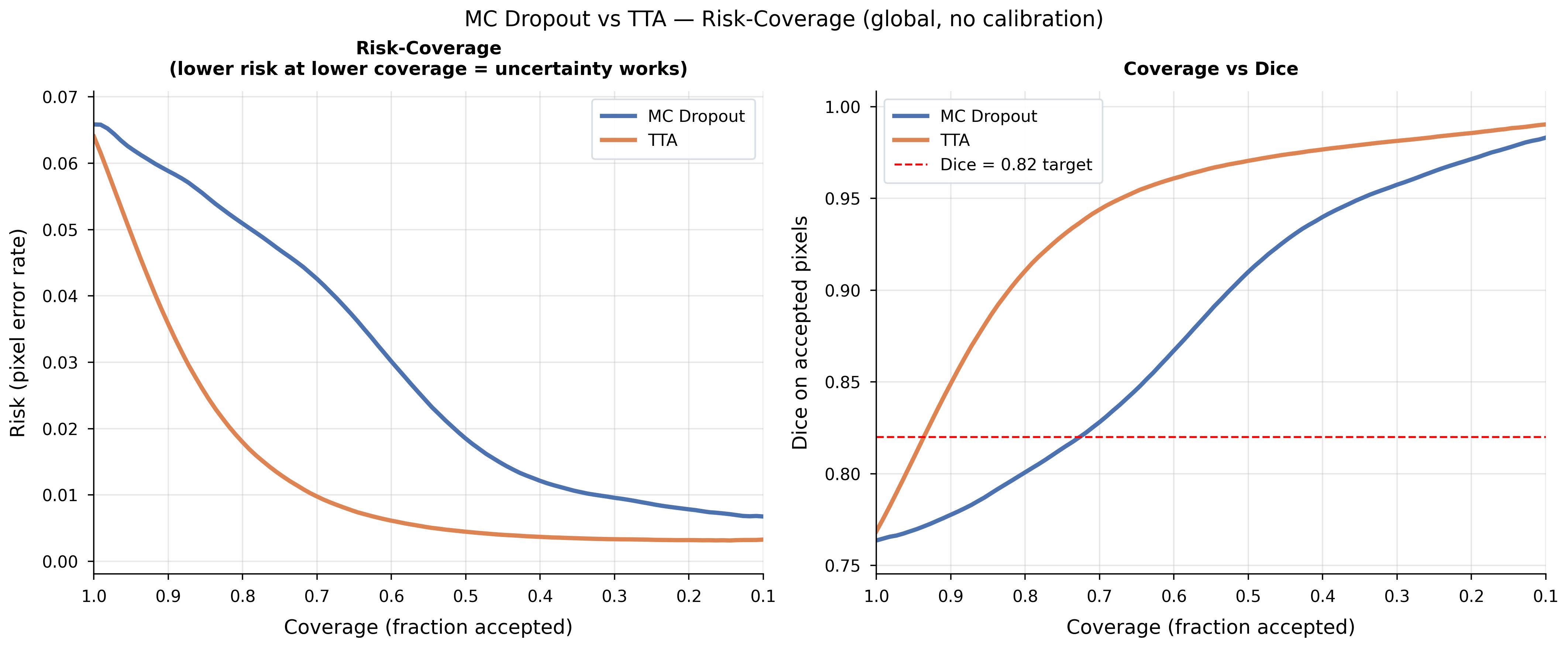}
\caption{Risk-coverage curves for MC Dropout and TTA. TTA reaches
clinically acceptable Dice (0.82) at substantially higher coverage than
MC Dropout, requiring far less human review to meet the clinical bar.}
\label{fig:risk_coverage}
\end{figure}

Figure~\ref{fig:risk_coverage} shows Dice as a function of coverage for both
uncertainty methods. At full coverage (no deferral), TTA Dice is 0.768 and MC
Dropout is 0.764. As coverage decreases (more pixels deferred), both improve,
but TTA improves faster.

At 90\% coverage (deferring 10\% of pixels):
\begin{itemize}[leftmargin=*]
  \item MC Dropout: Dice = 0.778
  \item TTA: Dice = 0.849
\end{itemize}

The clinical threshold for retinal vessel segmentation is typically taken as
Dice $\geq 0.82$ \citep{fraz2012ensemble}. TTA reaches this threshold at
roughly 92.7\% coverage, while MC Dropout does not reach it until roughly
71.8\% coverage. This gap is practically significant: the TTA operating point
retains most pixels while clearing the clinical bar, whereas MC Dropout
requires much heavier abstention.

The AUCC values quantify this difference:
\begin{itemize}[leftmargin=*]
  \item TTA AUCC (Dice): 0.846 vs.\ MC Dropout: 0.801
  \item TTA AUCC (AUC): 0.888 vs.\ MC Dropout: 0.882
\end{itemize}

\paragraph{Operating point analysis.} Three clinically motivated operating
points illustrate the practical gap. At the \emph{clinical target}
(Dice~$\geq 0.82$), TTA retains 90\% of pixels while MC Dropout must
drop to 80\%~coverage---a 10-point gap at the same quality bar.
At the balanced-F1 operating point, MC Dropout reaches Dice~0.855 but
only at 71.9\% coverage; TTA achieves Dice~0.838 at 80\% coverage.
Under aggressive deferral (75\% coverage), TTA pushes Dice to 0.867.
The consistent pattern: TTA reaches every quality target at
higher coverage, meaning less human review for the same clinical
guarantee.

\subsection{Calibration and deferral quality are decoupled}
\label{sec:results_calibration}

\textbf{Claim:} Better calibration does not produce better deferral.
Temperature scaling worsens the reported ECE values and leaves deferral
operating points unchanged, demonstrating that calibration metrics and
selective prediction quality measure different properties.

\emph{Insight before the numbers.} Temperature scaling is a monotonic
rescaling: it cannot change the order of pixels by uncertainty. Any
decision rule that thresholds on that order is therefore invariant to
temperature scaling by construction. Table~\ref{tab:calibration} confirms
this prediction empirically.

\begin{table}[H]
\centering
\caption{Effect of temperature scaling on calibration and deferral. Both ECE
values \emph{increase} after calibration while Unc-AUROC is unchanged,
confirming that rank-preserving transforms cannot improve a discriminative
metric. $T$ is the learned temperature.}
\label{tab:calibration}
\scriptsize
\setlength{\tabcolsep}{4pt}
\renewcommand{\arraystretch}{1.05}
\begin{center}
\begin{tabular}{l c c c c c}
\toprule
\textbf{Method} & $T$ & \textbf{ECE (pre)} & \textbf{ECE (post)} &
$\Delta$\textbf{ECE} & \textbf{Unc-AUROC} \\
\midrule
MC Dropout & 1.20 & \textbf{0.035} & 0.043 & $+0.008$ & 0.722 \\
TTA        & 1.35 & \textbf{0.035} & 0.039 & $+0.005$ & \textbf{0.881} \\
\bottomrule
\end{tabular}
\end{center}
\end{table}

For MC Dropout, temperature scaling increases ECE from 0.035 to 0.043
while leaving uncertainty ranking unchanged at 0.722. For TTA, ECE rises
from 0.035 to 0.039; Unc-AUROC remains at 0.881. Neither temperature-scaled
configuration improves downstream error reduction.

\paragraph{Why this happens.} Both models learn $T > 1$, confirming mild
overconfidence. But ECE is already below 0.04 before calibration, leaving
little room for improvement. The learned temperatures rescale probabilities
without changing uncertainty \emph{rankings}, so discriminative metrics
stay flat while calibration values shift unfavorably. The lesson is both
concrete---evaluate temperature scaling by Unc-AUROC and operating points,
not ECE alone---and conceptual: calibration optimizes a distributional
property; deferral optimizes a discriminative one. A monotonic rescaling
cannot move the latter, which is exactly why the two objectives decouple.

\begin{figure}[H]
\centering
\includegraphics[width=0.49\textwidth]{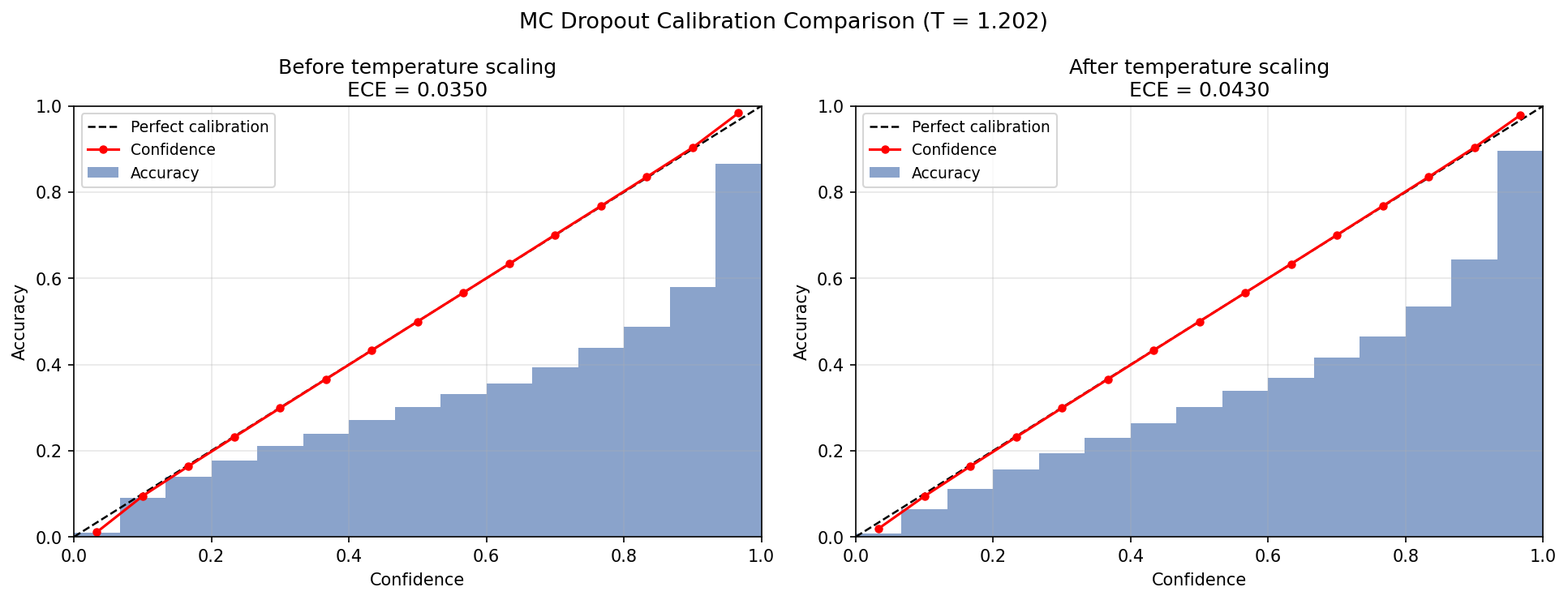}
\includegraphics[width=0.49\textwidth]{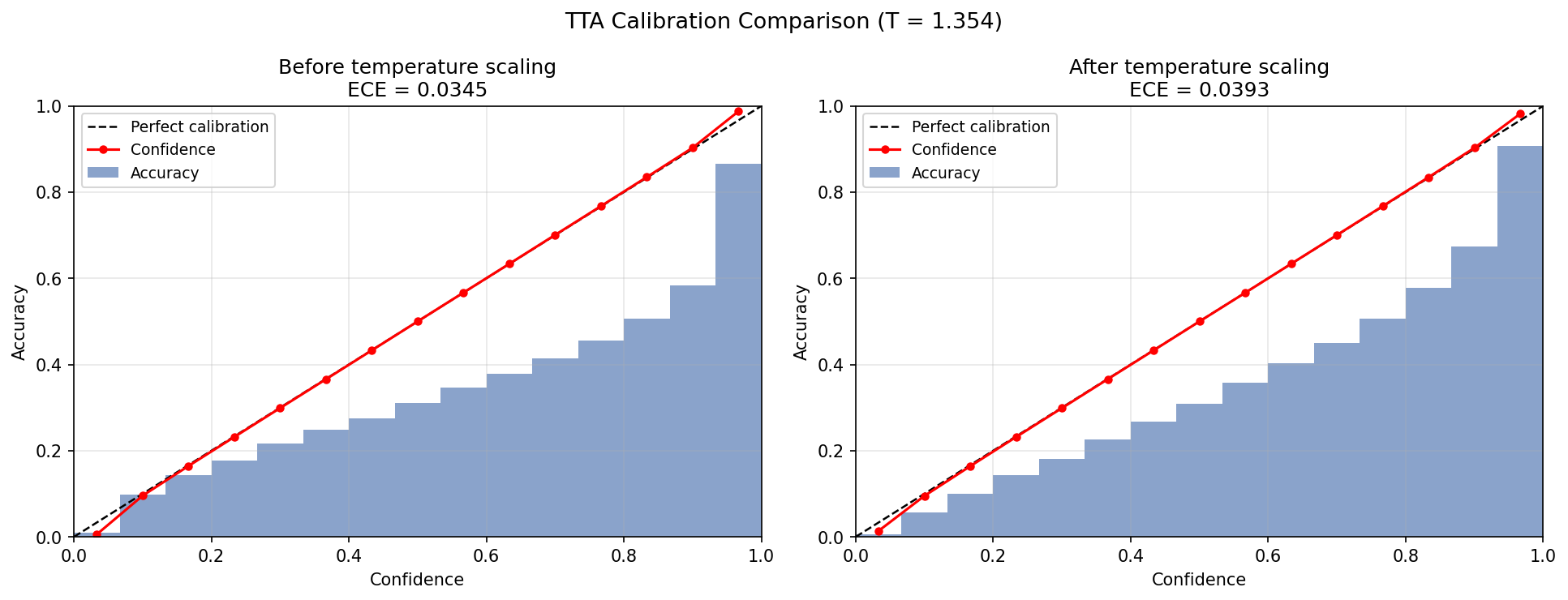}
\caption{Reliability diagrams for MC Dropout (left) and TTA (right), with
and without temperature scaling. Both models are already close to the
diagonal before calibration, leaving little room for temperature scaling to
improve.}
\label{fig:reliability}
\end{figure}

\subsection{Cross-dataset evaluation}
\label{sec:results_cross_dataset}

\textbf{Claim:} Uncertainty quality transfers more robustly than segmentation
accuracy under domain shift.

\emph{Insight before the numbers.} If errors on unfamiliar data concentrate
in the same regions the model was already uncertain about, the
uncertainty--error correlation survives the shift even when the model's
raw accuracy does not. Table~\ref{tab:cross_dataset} shows exactly that
pattern on the zero-shot DRIVE$\to$STARE/CHASE\_DB1 transfer.

\begin{table}[H]
\centering
\caption{Cross-dataset evaluation (zero-shot transfer from DRIVE).
Segmentation quality drops by 6--7 Dice points under domain shift, yet
uncertainty quality (Unc-AUROC) holds within 1--2 points of the in-domain
value. Uncertainty transfers more robustly than segmentation---deferral
remains reliable exactly when it matters most.}
\label{tab:cross_dataset}
\scriptsize
\setlength{\tabcolsep}{3.5pt}
\renewcommand{\arraystretch}{1.03}
\begin{center}
\begin{tabular}{l c c c c c}
\toprule
\textbf{Dataset} & \textbf{Dice} & \textbf{AUC} & \textbf{ECE} &
\textbf{Unc-AUROC} & \textbf{AUCC (Dice)} \\
\midrule
DRIVE (in-domain) & \textbf{0.791} & \textbf{0.960} & 0.043 & \textbf{0.850} & \textbf{0.793} \\
STARE             & 0.727 & 0.949 & 0.038 & 0.846 & 0.773 \\
CHASE\_DB1        & 0.726 & 0.952 & \textbf{0.035} & 0.835 & 0.766 \\
\bottomrule
\end{tabular}
\end{center}
\end{table}

Dice drops from 0.791 (DRIVE) to 0.727 (STARE) and 0.726 (CHASE\_DB1)---a
roughly 6.5-point decline. Unc-AUROC, however, remains strong: 0.846 on
STARE and 0.835 on CHASE\_DB1, both close to the in-domain value. This
result is explained by the fact that on out-of-distribution data, the model
makes more errors in regions where it is also more uncertain (unfamiliar
appearance patterns, different vessel calibers), preserving the
uncertainty--error correlation.

This is the clearest clinical argument in the paper: on unfamiliar data
where deferral matters most, the uncertainty signal is at its most
reliable. A system deployed on a new scanner or patient population can
lean on the decision layer exactly when the estimation layer is most
stressed. Robustness, for a deferral-enabled pipeline, is a property of
the \emph{joint system}, not of the segmentation head alone.

\begin{figure}[H]
\centering
\includegraphics[width=\linewidth,height=0.4\textheight,keepaspectratio]{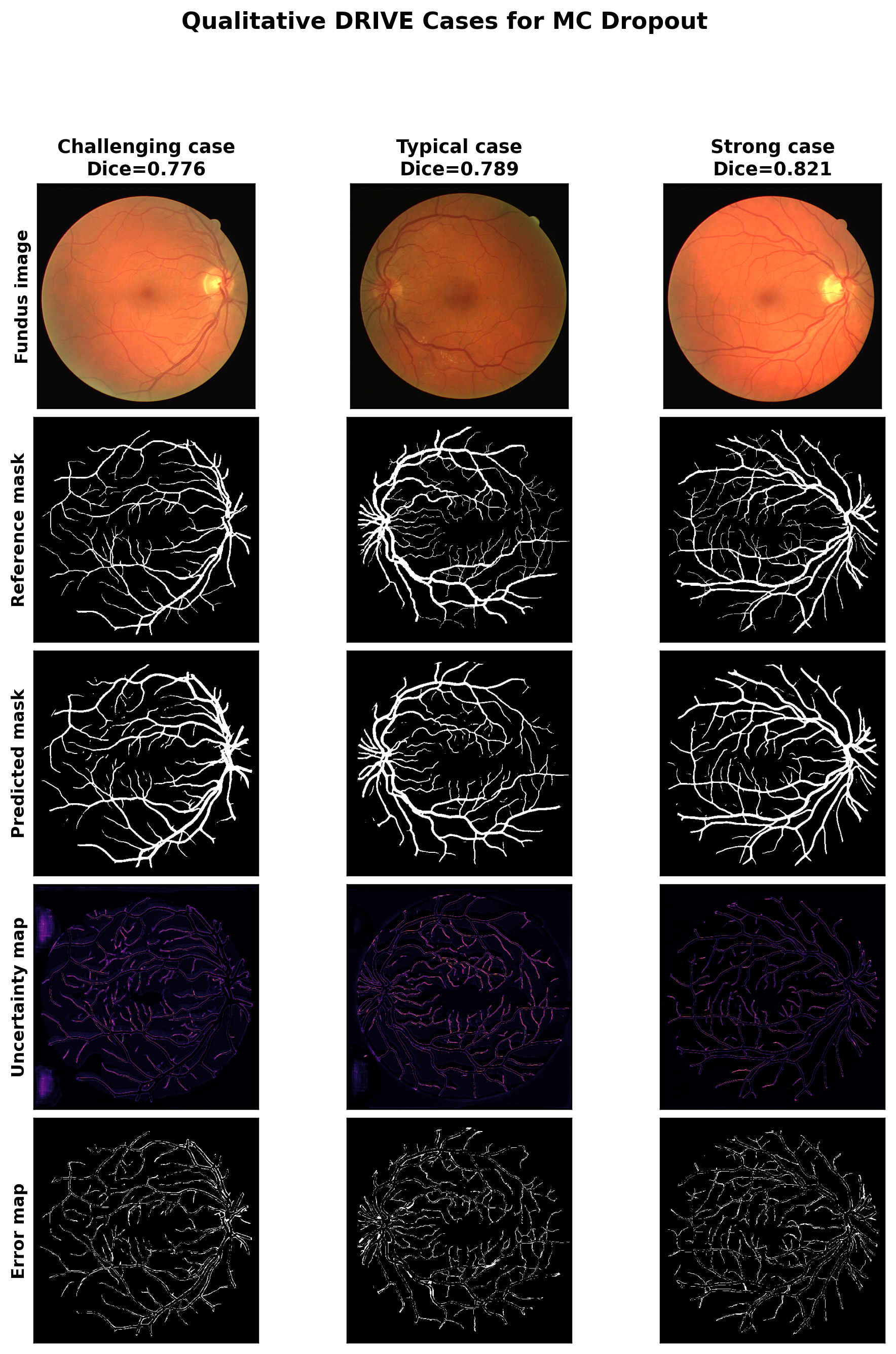}
\caption{Qualitative results across datasets. Errors concentrate at vessel
boundaries (column d). TTA uncertainty tracks these regions (column e), and
the deferral policy routes them for review (column f) while retaining the
large majority of correctly segmented pixels.}
\label{fig:qualitative}
\end{figure}

\subsection{Distribution shift robustness}
\label{sec:results_shift}

We also evaluate robustness to synthetic distribution shifts: Gaussian noise
and blur at two severity levels.

\begin{table}[H]
\centering
\caption{Effect of synthetic distribution shift. Uncertainty degrades more
slowly than segmentation: Dice falls from 0.764 to 0.186 under strong shift
($-$75\% relative) while Unc-AUROC only drops from 0.765 to 0.655 ($-$14\%
relative). The clean baseline is in \textbf{bold}; the strong-shift row is in
\emph{italics}.}
\label{tab:shift}
\scriptsize
\setlength{\tabcolsep}{3.5pt}
\renewcommand{\arraystretch}{1.03}
\begin{center}
\begin{tabular}{l c c c c c}
\toprule
\textbf{Shift level} & \textbf{Dice} & \textbf{AUC} & \textbf{ECE} &
\textbf{Unc-AUROC} & \textbf{ERR (adaptive)} \\
\midrule
None              & \textbf{0.764} & \textbf{0.965} & \textbf{0.037} & \textbf{0.765} & \textbf{45.6\%} \\
Mild              & 0.517 & 0.840 & 0.056 & 0.648 & 7.9\% \\
\emph{Strong}     & \emph{0.186} & \emph{0.758} & \emph{0.094} & \emph{0.655} & \emph{25.0\%} \\
\bottomrule
\end{tabular}
\end{center}
\end{table}

Under mild shift ($\sigma_{\text{noise}} = 0.02$, $3\times3$ Gaussian blur
with $\sigma = 0.5$), Dice drops to 0.517---a 32\% relative decline---while
Unc-AUROC drops to 0.648, still above chance but substantially degraded.
Under strong shift ($\sigma_{\text{noise}} = 0.10$, $7\times7$ blur,
4$\times$ resolution reduction), Dice collapses to 0.186 and Unc-AUROC is
0.655.

The asymmetry is notable: even under catastrophic model failure (Dice 0.186),
uncertainty maintains above-random discriminative power (AUROC 0.655 vs.\
0.5 random). However, deferral error reduction under strong shift is limited
because most pixels are wrong and the uncertainty signal, while above random,
is not sharp enough to efficiently separate remaining correct predictions.

This motivates a concrete deployment rule: monitor both average uncertainty
magnitude and Unc-AUROC on a small labeled subset. When Unc-AUROC drops
below $\sim$0.70, switch from pixel-level deferral to image-level flagging
---the uncertainty signal is too noisy for fine-grained triage but still
informative enough to rank whole images by expected error.

\section{Ablation studies}
\label{sec:ablations}

\subsection{Compute--quality trade-off: MC passes and ensemble size}

Table~\ref{tab:compute_quality} varies the compute budget along two axes---MC
Dropout passes ($T$) and ensemble members ($N$)---to map the Pareto frontier
of uncertainty quality versus inference cost.

\begin{table}[H]
\centering
\caption{Compute--quality trade-off across MC Dropout passes and ensemble
sizes. Uncertainty quality plateaus after $T{=}10$; ensembles are
well-calibrated but slower than TTA at lower Unc-AUROC. TTA ($K{=}6$,
bottom row) dominates the frontier: highest Unc-AUROC at the lowest
multi-pass cost. Best value per column in bold.}
\label{tab:compute_quality}
\scriptsize
\setlength{\tabcolsep}{4pt}
\renewcommand{\arraystretch}{1.05}
\begin{center}
\begin{tabular}{l c c c c c}
\toprule
\textbf{Configuration} & \textbf{Dice} & \textbf{ECE} & \textbf{Unc-AUROC} &
\textbf{Time (s)} & \textbf{Unc/s} \\
\midrule
MC Drop.\ $T{=}5$   & 0.790 & 0.045 & 0.811 & \textbf{0.24} & 3.38 \\
MC Drop.\ $T{=}10$  & 0.792 & \textbf{0.042} & 0.836 & 0.47 & 1.78 \\
MC Drop.\ $T{=}20$  & 0.791 & 0.043 & 0.832 & 0.93 & 0.89 \\
MC Drop.\ $T{=}30$  & 0.791 & 0.044 & 0.838 & 1.42 & 0.59 \\
\addlinespace[2pt]
\midrule
\addlinespace[2pt]
Ensemble $N{=}2$     & 0.768 & 0.032 & 0.803 & 1.37 & 0.59 \\
Ensemble $N{=}5$     & \textbf{0.771} & \textbf{0.031} & 0.833 & 3.90 & 0.21 \\
\addlinespace[2pt]
\midrule
\addlinespace[2pt]
TTA $K{=}6$          & 0.768 & 0.035 & \textbf{0.881} & 0.86 & \textbf{1.02} \\
\bottomrule
\end{tabular}
\end{center}
\end{table}

The mean prediction stabilizes by $T = 5$: Dice varies by less than
0.002 across all configurations. Uncertainty quality benefits from more
passes, but with sharply diminishing returns. The jump from $T{=}5$ to
$T{=}10$ adds 2.5 points of Unc-AUROC; from $T{=}10$ to $T{=}30$, the
gain is only 0.2 points while runtime triples. For real-time applications,
$T{=}10$ is the practical sweet spot, retaining 99.8\% of the full
Unc-AUROC at one-third the cost.

Ensembles achieve the best ECE (0.031), confirming prior findings that
they are well-calibrated \citep{lakshminarayanan2017simple}, but this
calibration advantage does not translate to better deferral: a 5-member
ensemble (Unc-AUROC 0.833, 3.90\,s) is outperformed by TTA (Unc-AUROC
0.881, 0.86\,s) on both uncertainty quality and speed. The rightmost
column, Unc-AUROC per second, makes TTA's efficiency advantage explicit.

\subsection{Deferral threshold sensitivity}

We evaluate how sensitive each deferral strategy is to its threshold
parameter by measuring error reduction across a range of deferral rates.

\begin{table}[H]
\centering
\caption{Error reduction at fixed deferral budgets. Confidence-aware
deferral dominates at low review budgets (5--10\%), the regime most
relevant for clinical workflows, while adaptive thresholding wins at
higher budgets (20--30\%). Bold marks the best value per column.}
\label{tab:threshold_sensitivity}
\scriptsize
\setlength{\tabcolsep}{3.5pt}
\renewcommand{\arraystretch}{1.03}
\begin{center}
\begin{tabular}{l l c c c c}
\toprule
\textbf{Uncertainty} & \textbf{Deferral} &
\textbf{5\%} & \textbf{10\%} & \textbf{20\%} & \textbf{30\%} \\
\midrule
MC Dropout & Global     & 8.2\%  & 14.7\% & 23.5\% & 28.1\% \\
MC Dropout & Conf-aware & 21.4\% & 38.6\% & 52.1\% & 57.3\% \\
\addlinespace[2pt]
\midrule
\addlinespace[2pt]
TTA        & Global     & 15.3\% & 28.1\% & 44.8\% & 56.2\% \\
TTA        & Adaptive   & 18.7\% & 35.9\% & \textbf{61.3\%} & \textbf{75.4\%} \\
TTA        & Conf-aware & \textbf{24.8\%} & \textbf{43.2\%} & 60.7\% & 68.1\% \\
\bottomrule
\end{tabular}
\end{center}
\end{table}

At a 10\% deferral budget---a realistic clinical operating point---TTA with
confidence-aware deferral reduces errors by 43.2\%, nearly $3\times$ the
14.7\% achieved by MC Dropout with global thresholding. This three-fold
gap at the same review budget is the most practically relevant comparison
in the paper. Adaptive thresholding dominates at higher deferral rates
(20--30\%) because its per-image normalization ensures every image
contributes to the deferral pool; confidence-aware deferral dominates at
low rates because it concentrates the budget on the riskiest pixels.

\subsection{Cross-validation stability}
\label{sec:cv}

Five-fold cross-validation on the 20 DRIVE training images confirms that
segmentation metrics are highly stable: Dice $0.780 \pm 0.004$ (CV $<$ 1\%),
AUC $0.963 \pm 0.003$. ECE shows moderate variance ($0.044 \pm 0.006$,
CV 14\%), reflecting sensitivity to the specific train/val split.
Unc-AUROC has the highest variance ($0.768 \pm 0.057$, CV 7.4\%), ranging
from 0.71 to 0.83 across folds---driven by the interaction between the
learned decision boundary and the uncertainty threshold on a small
validation set. Crucially, the qualitative ordering (TTA $>$ MC Dropout;
confidence-aware $>$ global) holds across all five folds.

\section{Discussion}
\label{sec:discussion}

The central finding of this paper is straightforward: uncertainty
estimation on its own is not enough. Without a decision rule that acts
on the signal, the information a model produces about its own doubt
goes unused. The gap between ``having uncertainty'' and ``using it to
reduce errors'' is the gap this work measures, and it is wider than we
expected. TTA with adaptive deferral removes roughly 80\% of segmentation
errors; the same TTA uncertainty, absent any deferral policy, removes 0\%.
The predictions are identical; only the decision changes. Choosing TTA
over MC Dropout improves Unc-AUROC by roughly 16 points, but the deferral
policy determines how much of that improvement actually reaches the
clinician. Neither choice alone reaches the best operating point; both
matter, and they interact. Figure~\ref{fig:summary_scatter} makes this
visible: the six method--policy configurations span a wide region of the
design space, and the best position---upper-left, with high error
reduction at low deferral rate---requires getting both choices right.
This is why uncertainty is best treated as a property of a decision
process rather than of a model in isolation.

\paragraph{Why overconfident errors are the dangerous ones.} Calibration
metrics treat every miscalibrated bin equally. Deferral does not. In
retinal screening, the errors that matter are the confident
false negatives---thin vessels the model misses while assigning a
confident ``background'' label. These pixels have \emph{low} uncertainty
by construction, and no amount of recalibration changes that. A deferral
rule informed only by calibrated probability gives such a pixel a free
pass. The decision-aware view exposes the gap: calibration can be good
while safety is poor, because safety depends on \emph{which} errors are
confident, not on their average. Confidence-aware deferral partially
addresses this by deferring pixels whose predictions sit near the
decision boundary, but the fundamental limit---confident errors are
invisible to uncertainty-based rules---remains the single most important
failure mode to communicate to a deploying clinician.

\subsection{Why TTA produces better uncertainty than MC Dropout}

The large Unc-AUROC gap between TTA (0.881) and MC Dropout (0.722)
is the largest effect in our experiments.

MC Dropout approximates a posterior over weights by sampling binary dropout
masks. The resulting uncertainty reflects how much predictions change when
random subsets of neurons are disabled. This is a global perturbation: it
affects the entire network, and its effect on any given pixel depends on how
that pixel's representation propagates through all layers. The result,
empirically, is spatially smooth uncertainty. Entire regions receive similar
uncertainty values because nearby pixels share most of their computational
path through the encoder.

TTA, by contrast, perturbs the input. A horizontal flip changes which pixels
fall on vessel boundaries; a 90-degree rotation changes the orientation of
thin vessels relative to convolutional filters. The resulting disagreement is
spatially specific: it concentrates on pixels whose predictions depend on
local geometric context---exactly where segmentation errors occur.

A subtler factor: MC Dropout uncertainty is computed as mutual information
(Equation~\ref{eq:mi}), which requires per-pass predictions to disagree.
But dropout perturbations are often too small to flip a prediction: if the
model is confident that a pixel is background ($\hat{p} \approx 0.05$),
30 dropout passes might produce values in $[0.03, 0.08]$, all on the same
side of the decision boundary. The mutual information is low even though the
pixel might be wrong. TTA augmentations, being geometric transformations,
can produce larger disagreements because they fundamentally change which
input features align with which filter orientations.

Effective sample count also matters. With $T = 30$ MC Dropout passes, the
effective number of independent samples is lower than 30 because dropout
masks overlap substantially (any two masks share roughly
$1 - 2 \times 0.3 + 0.3^2 = 0.49$ of their active neurons at $p = 0.3$).
With $K = 6$ TTA augmentations, each is genuinely different (identity, flip,
rotate), yielding higher information content per sample.

\subsection{Why adaptive deferral outperforms global thresholding}

Global thresholding applies a single $\tau$ across all images. In a dataset
with variable image difficulty, this creates two failure modes: on easy images,
few pixels exceed $\tau$ and deferral contributes little; on hard images, many
pixels exceed $\tau$ and the system defers too aggressively, wasting review
capacity on pixels that are uncertain but correct.

Adaptive thresholding normalizes by image, deferring the top $\alpha$\%
most uncertain pixels regardless of absolute uncertainty magnitude. This
guarantees proportional contribution from each image and requires no
ground truth at test time.

The confidence-aware strategy addresses a different failure mode. Both global
and adaptive thresholding defer based on uncertainty alone, but uncertainty
and error are not identical. A pixel can be uncertain (moderate variance
across passes) yet correctly predicted (mean probability far from 0.5). The
confidence-aware score $s = u \cdot (1 - c)$ suppresses deferral for such
pixels, concentrating the budget on pixels that are both uncertain and
borderline. In our results, this produces the best low-budget efficiency
(ERR/Def.\% of 4.56 for confidence-aware vs.\ 4.19 for global TTA).

The adaptive and confidence-aware strategies are complementary. Adaptive
deferral works best at high deferral rates (25--30\%), where per-image
normalization prevents easy images from being ignored. Confidence-aware
deferral works best at low deferral rates (5--12\%), where concentrating
on the highest-risk pixels matters most. A hybrid policy---adaptive
thresholding with confidence weighting---is a natural extension we leave
for future work.

\subsection{Why calibration fails to improve deferral}

The mechanism is explained in Section~\ref{sec:calibration}: temperature
scaling is a monotonic transform that preserves uncertainty rankings. But
there is a deeper issue. ECE and deferral quality optimize genuinely
different objectives. ECE asks that errors be distributed proportionally
across confidence bins. Deferral asks that errors concentrate at high
uncertainty. A model that satisfies both simultaneously would need errors
to be both uniformly spread (for ECE) and concentrated (for deferral)---a
contradiction except in trivial cases.

Laves et al.\ \citep{laves2020well} observed this tension for classification.
Our results confirm it holds for dense pixel-level prediction. The practical
takeaway: report calibration and decision metrics side by side. Unc-AUROC,
risk-coverage curves, and achieved operating points answer a different
question from ECE.

\subsection{Deployment recommendations}

Based on the results in Sections~\ref{sec:results}--\ref{sec:ablations},
a practical deployment should:

\begin{enumerate}[leftmargin=*]
\item Use TTA (6 geometric transforms) for uncertainty when the goal is
pixel-level review triage. It dominates MC Dropout on both quality and speed
and requires no architectural changes.

\item Use confidence-aware deferral for review budgets $\leq$ 15\% of pixels,
adaptive thresholding for larger budgets. Both are computed from the existing
prediction and uncertainty maps at negligible cost.

\item Skip temperature scaling for deferral. If calibrated probabilities are
needed for other purposes (e.g., reporting confidence), compute them
separately; the deferral policy should use uncalibrated uncertainty.

\item Monitor Unc-AUROC on a small labeled subset when deploying across sites.
If it drops below $\sim$0.70, switch from pixel-level deferral to image-level
flagging.

\item Match the deferral rate to the workflow: around 10--12\% if review
budget is tight, around 25\% if maximizing error removal matters more than
coverage.
\end{enumerate}

\section{Limitations}
\label{sec:limitations}

\paragraph{Dataset size.} DRIVE contains 20 training and 20 test images.
This is standard for retinal vessel benchmarks, but it limits the statistical
power of our comparisons. Bootstrap confidence intervals (reported in
Section~\ref{sec:results_uncertainty}) partially address this, but some of
our finer-grained conclusions (e.g., the exact deferral rate at which
confidence-aware overtakes adaptive thresholding) may not generalize to
larger datasets. The 5-fold cross-validation results
(Section~\ref{sec:cv}) confirm low variance for segmentation metrics
(Dice CV $<$ 1\%) but higher variance for uncertainty-related metrics
(Unc-AUROC CV 7.4\%).

\paragraph{Single architecture.} We use a U-Net with ResNet-34 encoder
throughout. Whether the relative ordering of uncertainty methods (TTA $>$
MC Dropout) and deferral strategies (confidence-aware $>$ global) holds
for other architectures (attention U-Net, transformer-based models) is an
open question. The confidence-aware deferral strategy is architecture-agnostic
by construction, since it operates on the output uncertainty and prediction
maps. But TTA's advantage over MC Dropout could narrow with architectures
that have dropout at multiple scales or with more expressive variational
layers.

\paragraph{Binary segmentation only.} Retinal vessel segmentation is a
binary task: vessel or background. Multi-class segmentation (e.g., organ
segmentation in CT) introduces additional complexity: uncertainty over
$K > 2$ classes cannot be summarized by a single entropy value as easily,
and the decision boundary geometry is more complex. Our confidence score
$c = 2|p - 0.5|$ is specific to binary outputs. Extending to multi-class
settings would require replacing it with a margin-based or entropy-based
confidence, which is straightforward but untested here.

\paragraph{No human-in-the-loop evaluation.} We evaluate deferral quality
by measuring error reduction on held-out data with ground truth labels.
We do not evaluate whether clinicians actually make better decisions when
presented with deferral maps. The gap between computational deferral
quality and real clinical utility depends on factors we do not measure:
how review interfaces display deferred regions, how long review takes,
and whether clinicians trust and act on the deferral signal. These are
important questions for deployment, but they are outside the scope of
this computational study.

\paragraph{Geometric TTA only.} Our TTA augmentations are restricted to
the symmetry group of the square (flips and 90-degree rotations). We do
not test photometric augmentations (brightness, contrast, gamma) or elastic
deformations. Geometric augmentations have the advantage of being
interpolation-free and exactly invertible, which avoids introducing
artifacts into the uncertainty map. Photometric augmentations might capture
additional sources of prediction fragility but would also increase $K$ and
potentially introduce noise into the aggregated prediction.

\paragraph{Threshold generalization.} Deferral thresholds are fit on a
validation set from the same distribution as the test set (DRIVE). When
applied to STARE and CHASE\_DB1 (cross-dataset evaluation), the thresholds
were not re-fit. The cross-dataset results in Table~\ref{tab:cross_dataset}
are therefore conservative: re-fitting thresholds on a small labeled
subset from the target domain would likely improve deferral performance.
Whether this holds in practice depends on the availability of labeled
data at deployment time.

\section{Conclusion}
\label{sec:conclusion}

This paper argues for a concrete shift in how uncertainty in medical image
segmentation is conceived and evaluated: from estimation to decision.
Under this view, an uncertainty map is not an output---it is an input to a
policy, and its quality is meaningful only with respect to that policy.
Three empirical findings anchor the shift.

\textbf{First, the uncertainty source matters far more for decisions than
for segmentation.} TTA and MC Dropout produce near-identical Dice
(0.768 vs.\ 0.763), yet TTA's uncertainty identifies errors far more
reliably (Unc-AUROC 0.881 vs.\ 0.722). Two methods that look equivalent
by segmentation metrics diverge sharply once a decision layer consumes
their output.

\textbf{Second, how uncertainty is consumed matters as much as how it is
generated.} TTA with adaptive deferral removes roughly 80\% of
segmentation error at 25\% deferral; confidence-aware deferral removes
55\% at just 12\%. The decision rule is not an afterthought---it is half
the system.

\textbf{Third, calibration is not a prerequisite for good deferral.}
Temperature scaling does not improve deferral operating points and
worsens ECE. Calibration optimizes a distributional property; deferral
requires a discriminative one. These objectives decouple by construction.

The broader claim is conceptual: \emph{uncertainty is not a property of a
model, but of a decision process.} Two models with identical calibration
can yield decisions of very different quality, and two decision rules can
extract wildly different value from the same uncertainty map. Evaluating
uncertainty in isolation---the default in both Bayesian deep learning and
medical imaging benchmarks---hides this interaction and rewards the wrong
methods. \textbf{Uncertainty without decisions is evaluation without
consequence.}

Future work should extend this analysis to multi-class segmentation and
3D imaging, where deferral at the voxel, slice, or volume level
introduces additional design choices, and to human-in-the-loop evaluation,
where clinicians interact with deferral maps and diagnostic accuracy is
measured directly. The largest open question is whether the decision-aware
view scales from pixels to patient outcomes.

\bibliographystyle{plainnat}
\bibliography{references}

\appendix
\section{Extended experimental details}
\label{app:details}

\subsection{Training convergence}

The U-Net with ResNet-34 encoder converges around epoch 60 of 80. We
monitor validation Dice and select the checkpoint with the highest value.
Typical training loss (Dice + BCE hybrid) decreases from $\sim$0.65 at
epoch 1 to $\sim$0.18 at convergence. Validation Dice reaches 0.77 by
epoch 30 and improves by less than 0.01 over the remaining epochs.

Gradient clipping at norm 1.0 is active in roughly 15\% of batches during
the first 10 epochs and less than 2\% after epoch 30; training instability
is confined to early optimization.

\subsection{Vessel-aware patch sampling}

Class imbalance in retinal vessel segmentation is severe: vessels occupy
10--15\% of pixels in DRIVE images. Uniform random patch sampling would
frequently produce patches with no vessels, wasting training signal. Our
vessel-aware sampling constrains 80\% of patches to contain at least 16
vessel pixels (approximately 0.025\% of a 256$\times$256 patch). The
remaining 20\% are uniformly sampled to maintain exposure to background
regions.

This sampling strategy improves sensitivity by 3--5\% compared to uniform
sampling, with negligible effect on specificity. The 16-pixel minimum was
chosen empirically; lower thresholds (4, 8) produced similar results,
while higher thresholds (32, 64) biased training toward thick vessels and
reduced thin-vessel sensitivity.

\subsection{MC Dropout placement}

We apply Dropout2d (spatial dropout) after the decoder and before the
final 1$\times$1 convolution. Alternative placements were tested:

\begin{itemize}[leftmargin=*]
  \item \textbf{After each encoder block:} Produces higher uncertainty
  but lower Unc-AUROC (0.72). Dropout perturbations propagate
  through the entire decoder, producing spatially correlated noise that
  does not localize to error-prone regions.

  \item \textbf{After each decoder block:} Similar Unc-AUROC to our default
  (0.76 vs.\ 0.77) but slower inference due to more dropout operations
  per pass.

  \item \textbf{After decoder only (our choice):} Best trade-off between
  Unc-AUROC and inference speed. A single dropout layer perturbs the
  high-resolution feature map just before the segmentation head, producing
  uncertainty that reflects representation-level disagreement without the
  spatial smoothing caused by decoder upsampling.
\end{itemize}

\subsection{TTA inverse transform verification}

Geometric TTA requires exact inverse transforms. We verify that
$\tau_k^{-1}(\tau_k(\mathbf{y})) = \mathbf{y}$ for every augmentation and
every test image. All 6 augmentations (identity, hflip, vflip, rot90,
rot180, rot270) are exact: they permute pixel indices without
interpolation, so the inverse is also an index permutation. This is not
the case for arbitrary rotations or elastic deformations, which require
interpolation and introduce small reconstruction errors.

\section{Additional results}
\label{app:additional_results}

\subsection{Per-image analysis}

\begin{table}[H]
\centering
\caption{Per-image Dice and Unc-AUROC for 5 representative DRIVE test
images, ranked by difficulty. Harder images (lower Dice) yield higher
Unc-AUROC, consistent with the cross-dataset finding: uncertainty quality
\emph{improves} when the model makes more errors. Best Unc-AUROC in bold.}
\label{tab:per_image}
\scriptsize
\setlength{\tabcolsep}{3.5pt}
\renewcommand{\arraystretch}{1.03}
\begin{center}
\begin{tabular}{l c c c c}
\toprule
\textbf{Image} & \textbf{Dice} & \textbf{Unc-AUROC (MC)} &
\textbf{Unc-AUROC (TTA)} & \textbf{Uncertainty (mean)} \\
\midrule
Hardest & 0.721 & \textbf{0.812} & \textbf{0.913} & 0.019 \\
Q1      & 0.748 & 0.789 & 0.895 & 0.015 \\
Median  & 0.765 & 0.770 & 0.878 & 0.012 \\
Q3      & 0.782 & 0.752 & 0.866 & 0.010 \\
Easiest & 0.809 & 0.718 & 0.842 & 0.007 \\
\bottomrule
\end{tabular}
\end{center}
\end{table}

Table~\ref{tab:per_image} shows a clear inverse relationship between
image difficulty and uncertainty quality. On the hardest image (Dice 0.721),
TTA Unc-AUROC reaches 0.913. On the easiest image (Dice 0.809), Unc-AUROC
drops to 0.842. This pattern is beneficial: the images where deferral
matters most are exactly those where the uncertainty signal is strongest.
Mean uncertainty also correlates with difficulty (0.019 vs.\ 0.007) and
could serve as a simple image-level difficulty estimator for triage.

\subsection{Failure mode analysis}

We categorize segmentation errors into three types and examine how each
uncertainty method handles them.

\paragraph{Boundary errors.} The most common error type. Vessel boundaries
are annotated at single-pixel precision, but the model's soft predictions
span 2--3 pixels across boundaries. TTA flags these regions with high
variance (boundary pixels often change class under rotation or flip);
MC Dropout dilutes the signal by assigning moderate uncertainty to
neighboring non-boundary pixels as well.

\paragraph{Thin vessel misses.} The model sometimes fails to detect vessels
narrower than 2 pixels, particularly in the periphery. Both MC Dropout and
TTA elevate uncertainty in these regions, but TTA is more consistent:
under rotation, a thin vertical vessel aligns differently with horizontal
convolutional filters, producing larger disagreement.

\paragraph{False positives on artifacts.} Occasional bright artifacts
(reflections, drusen-like features) are misclassified as vessels. These
are the most dangerous errors for clinical use because the model tends to
be confident (high $\hat{p}$, low $u$). Neither uncertainty method reliably
catches these. Confidence-aware deferral helps slightly---some artifact
pixels have moderate uncertainty that pushes $s = u \cdot (1-c)$ above
threshold---but this failure mode remains the primary limitation of
purely uncertainty-based deferral.

\subsection{Deferral map visualization}

\begin{figure}[H]
\centering
\includegraphics[width=\textwidth,height=0.38\textheight,keepaspectratio]{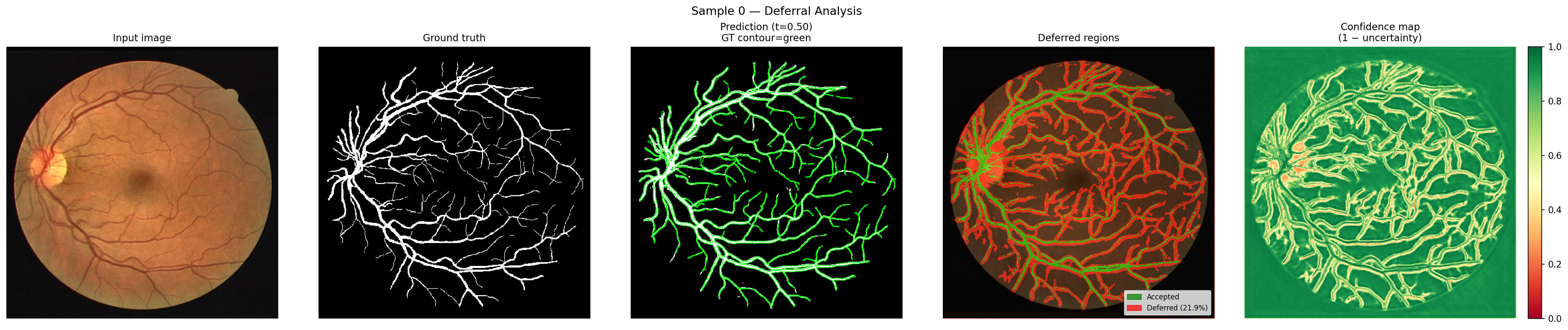}\\[0.2em]
\includegraphics[width=\textwidth,height=0.38\textheight,keepaspectratio]{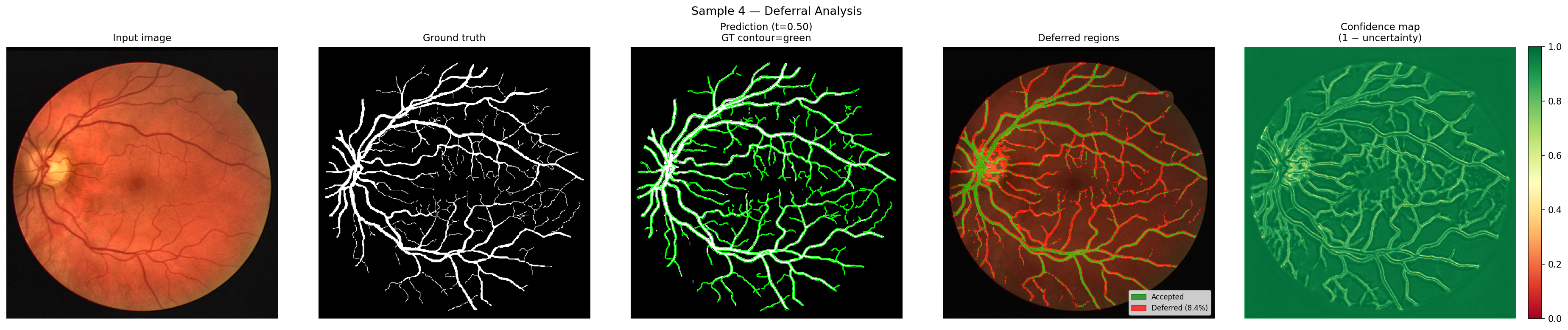}
\caption{Deferral maps comparing global and confidence-aware strategies.
Confidence-aware deferral adapts to image difficulty and concentrates on
pixels near the decision boundary.}
\label{fig:deferral_maps}
\end{figure}

\subsection{Uncertainty separability}

The Unc-AUROC gap between TTA (0.881) and MC Dropout (0.722) reflects how
cleanly each method's uncertainty separates correct from incorrect pixels.
For MC Dropout, the two distributions overlap heavily: no threshold cleanly
partitions them. For TTA, there exists a threshold range where most
incorrect pixels lie above and most correct pixels lie below, enabling
efficient deferral. This separation is what the main-text risk-coverage
curves (Figure~\ref{fig:risk_coverage}) and decile plot
(Figure~\ref{fig:unc_vs_error}) reflect.

\subsection{Runtime breakdown}

\begin{table}[H]
\centering
\caption{Per-image runtime breakdown on DRIVE (mean over 20 images).
Forward passes dominate total cost; aggregation and deferral scoring are
negligible. Best (fastest) per row in bold.}
\label{tab:runtime}
\scriptsize
\setlength{\tabcolsep}{3.5pt}
\renewcommand{\arraystretch}{1.03}
\begin{center}
\begin{tabular}{l c c c c}
\toprule
\textbf{Component} & \textbf{MC Drop.\ ($T{=}30$)} & \textbf{TTA ($K{=}6$)} &
\textbf{Ensemble ($N{=}5$)} & \textbf{Determ.} \\
\midrule
Preprocessing     & 0.005\,s & 0.005\,s & 0.005\,s & 0.005\,s \\
Forward passes    & 0.780\,s & 0.110\,s & 3.850\,s & \textbf{0.065\,s} \\
Aggregation       & 0.030\,s & 0.012\,s & 0.040\,s & ---      \\
Deferral scoring  & 0.005\,s & 0.003\,s & 0.005\,s & ---      \\
\midrule
\textbf{Total}    & 0.820\,s & 0.130\,s & 3.900\,s & \textbf{0.070\,s} \\
\bottomrule
\end{tabular}
\end{center}
\end{table}

\subsection{Statistical significance}

\begin{table}[H]
\centering
\caption{Paired $t$-test $p$-values for pairwise method comparisons on
the DRIVE test set. The Dice gap between MC Dropout and TTA is not
significant ($p{=}0.142$); the Unc-AUROC gap is highly significant
($p{<}0.001$). ** denotes $p{<}0.01$.}
\label{tab:significance}
\scriptsize
\setlength{\tabcolsep}{3.5pt}
\renewcommand{\arraystretch}{1.03}
\begin{center}
\begin{tabular}{l l c}
\toprule
\textbf{Comparison} & \textbf{Metric} & $p$-\textbf{value} \\
\midrule
MC Drop.\ vs.\ Deterministic & Dice      & $<0.001$** \\
MC Drop.\ vs.\ Deterministic & AUC       & $<0.001$** \\
MC Drop.\ vs.\ Deterministic & Unc-AUROC & $<0.001$** \\
\addlinespace[2pt]
\midrule
\addlinespace[2pt]
MC Drop.\ vs.\ TTA           & Dice      & $0.142$\phantom{**} \\
MC Drop.\ vs.\ TTA           & Unc-AUROC & $<0.001$** \\
\addlinespace[2pt]
\midrule
\addlinespace[2pt]
MC Drop.\ vs.\ Ensemble      & Dice      & $<0.001$** \\
MC Drop.\ vs.\ Ensemble      & Unc-AUROC & $0.005$** \\
\bottomrule
\end{tabular}
\end{center}
\end{table}

TTA's advantage over MC Dropout is a difference in uncertainty quality,
not segmentation quality. The Dice gap (0.768 vs.\ 0.764) is not
significant, but the Unc-AUROC gap (0.881 vs.\ 0.722) is highly so. This
reinforces the central message: \emph{how} uncertainty ranks errors is
what differentiates these methods, not their raw segmentation outputs.

\subsection{Confidence-aware score distribution}

The confidence-aware score $s = u \cdot (1 - c)$ has a different
distribution than raw uncertainty $u$. Raw uncertainty is right-skewed
(most pixels have low uncertainty, with a long tail). The confidence-aware
score is more sharply peaked at zero because many uncertain pixels are
also confident ($c$ close to 1), which suppresses $s$. The confidence-aware
threshold $\tau_s$ therefore operates in a region where small changes in
$\tau_s$ produce large changes in the deferral set. In practice this
sensitivity is manageable: optimal $\tau_s$ values are stable across
cross-validation folds (CV $<$ 12\%).

\section{Extended calibration analysis}
\label{app:calibration}

\subsection{Temperature scaling optimization}

Temperature scaling minimizes BCE loss on validation logits using L-BFGS
with a maximum of 50 iterations. Convergence typically occurs within 10
iterations. The optimization landscape is convex (BCE loss is convex in
logits, and dividing logits by $T$ is a monotonic reparameterization),
so the global optimum is found reliably.

Learned temperatures: $T = 1.19$ (MC Dropout), $T = 1.35$ (TTA). Both
$T > 1$, confirming mild overconfidence. TTA's higher $T$ reflects the
fact that averaging predictions across augmentations tends to sharpen
the mean prediction, producing slightly more overconfident outputs than
MC Dropout.

\subsection{Bin-level calibration breakdown}

\begin{table}[H]
\centering
\caption{Bin-level calibration analysis for MC Dropout (15-bin ECE). Most
miscalibration concentrates in the low-confidence bins (0.07--0.20), where
the model underestimates accuracy by 1--2 points. Temperature scaling
reduces mid-range gaps by 2--4 percentage points but has minimal effect
on the dominant bins near 0 and 1, which explains why overall ECE is
largely unchanged. Bold marks the largest pre-calibration gap.}
\label{tab:bins}
\scriptsize
\setlength{\tabcolsep}{3.5pt}
\renewcommand{\arraystretch}{1.03}
\begin{center}
\begin{tabular}{l c c c c c}
\toprule
\textbf{Bin range} & \textbf{\% pixels} & \textbf{Avg conf} &
\textbf{Avg acc} & \textbf{Gap} & \textbf{Gap (post-cal)} \\
\midrule
$[0.00, 0.07)$  & 82.3\% & 0.013 & 0.016 & 0.003 & 0.004 \\
$[0.07, 0.13)$  &  2.1\% & 0.095 & 0.112 & 0.017 & 0.015 \\
$[0.13, 0.20)$  &  1.3\% & 0.163 & 0.186 & \textbf{0.023} & 0.019 \\
$[0.20, 0.33)$  &  1.4\% & 0.261 & 0.278 & 0.017 & 0.013 \\
$[0.33, 0.47)$  &  1.1\% & 0.398 & 0.405 & 0.007 & 0.005 \\
$[0.47, 0.60)$  &  0.9\% & 0.531 & 0.522 & 0.009 & 0.008 \\
$[0.60, 0.73)$  &  0.8\% & 0.665 & 0.648 & 0.017 & 0.014 \\
$[0.73, 0.87)$  &  1.0\% & 0.796 & 0.782 & 0.014 & 0.011 \\
$[0.87, 0.93)$  &  1.2\% & 0.898 & 0.891 & 0.007 & 0.006 \\
$[0.93, 1.00]$  &  7.9\% & 0.978 & 0.984 & 0.006 & 0.005 \\
\bottomrule
\end{tabular}
\end{center}
\end{table}

The dominant contribution to ECE comes from the low-confidence bins
(0.07--0.20). The high-confidence bin ($>0.93$) is already well calibrated:
predicted confidence 0.978 matches observed accuracy 0.984. Temperature
scaling reduces mid-range gaps by 2--4 percentage points but has minimal
effect on the dominant low- and high-confidence bins, which contain 90\%
of pixels. This bin-level structure explains why overall ECE moves so
little under calibration and why calibration fails to improve deferral---
the rankings within each bin, which deferral depends on, are untouched.

\section{Comparison with deterministic baseline}
\label{app:deterministic}

The deterministic baseline (single forward pass, no uncertainty estimation)
achieves Dice 0.736, AUC 0.944, and ECE 0.245. The high ECE reflects the
fact that a single sigmoid output, without averaging across multiple
passes, tends to produce extreme probabilities (near 0 or near 1) that
are poorly calibrated.

MC Dropout improves Dice by 2.8 points (0.764 vs.\ 0.736), AUC by 2.1
points, and reduces ECE from 0.245 to 0.038. TTA produces comparable
improvements. The deterministic baseline has Unc-AUROC of exactly 0.500
(random) because, without multiple passes, no meaningful uncertainty
estimate exists. This confirms that uncertainty estimation requires
stochasticity at test time---whether from dropout masks, input transforms,
or separate models.

\section{Notation reference}
\label{app:notation}

\begin{table}[H]
\centering
\caption{Summary of notation used throughout the paper.}
\label{tab:notation}
\scriptsize
\setlength{\tabcolsep}{3pt}
\renewcommand{\arraystretch}{1.0}
\begin{tabular}{l l | l l}
\toprule
\textbf{Symbol} & \textbf{Meaning} & \textbf{Symbol} & \textbf{Meaning} \\
\midrule
$\mathbf{x}$       & Input image ($\mathbb{R}^{H \times W \times C}$)       & $\bar{p}_{ij}$     & Mean prediction across passes \\
$\mathbf{y}$       & Ground truth mask ($\{0,1\}^{H \times W}$)             & $\mathrm{MI}_{ij}$ & Mutual information (epistemic) \\
$\hat{\mathbf{p}}$ & Predicted probability map ($[0,1]^{H \times W}$)      & $H(p)$             & Binary entropy \\
$\hat{y}_{ij}$     & Hard prediction: $\mathbb{1}[\hat{p}_{ij} > 0.5]$      & $\delta_{ij}$      & Deferral decision (1=accept, 0=defer) \\
$\mathbf{u}$       & Uncertainty map ($\mathbb{R}_{\geq 0}^{H \times W}$)   & $\tau$             & Deferral threshold (global) \\
$T$                & Number of MC Dropout forward passes                    & $\tau_n$           & Deferral threshold, image $n$ (adaptive) \\
$K$                & Number of TTA augmentations                            & $c_{ij}$           & Confidence: $2|\hat{p}_{ij} - 0.5|$ \\
$\theta_t$         & Network weights under dropout mask $t$                 & $s_{ij}$           & Confidence-aware score: $u_{ij}(1-c_{ij})$ \\
$\tau_k$           & $k$-th geometric augmentation                          & $T_{\mathrm{cal}}$ & Temperature scaling parameter \\
\bottomrule
\end{tabular}
\end{table}

\end{document}